\documentclass{ieeeaccess}[preprint]
\usepackage{cite}
\usepackage{amsmath,amssymb,amsfonts}
\usepackage{algorithmic}
\usepackage{graphicx}
\usepackage{textcomp}

\usepackage{bm}
\makeatletter
\AtBeginDocument{\DeclareMathVersion{bold}
\SetSymbolFont{operators}{bold}{T1}{times}{b}{n}
\SetSymbolFont{NewLetters}{bold}{T1}{times}{b}{it}
\SetMathAlphabet{\mathrm}{bold}{T1}{times}{b}{n}
\SetMathAlphabet{\mathit}{bold}{T1}{times}{b}{it}
\SetMathAlphabet{\mathbf}{bold}{T1}{times}{b}{n}
\SetMathAlphabet{\mathtt}{bold}{OT1}{pcr}{b}{n}
\SetSymbolFont{symbols}{bold}{OMS}{cmsy}{b}{n}
\renewcommand\boldmath{\@nomath\boldmath\mathversion{bold}}}
\makeatother

\def\BibTeX{{\rm B\kern-.05em{\sc i\kern-.025em b}\kern-.08em
    T\kern-.1667em\lower.7ex\hbox{E}\kern-.125emX}}
    
\usepackage[hyphens]{url}
\usepackage[
    colorlinks=true,
    allcolors=accessblue,
    breaklinks=true
]{hyperref}
\usepackage{multirow}
\usepackage[absolute]{textpos}
\begin{document}
\history{Date of publication xxxx 00, 0000, date of current version xxxx 00, 0000.}
\doi{10.1109/ACCESS.2026.3678816}

\title{LeLaR: The First In-Orbit Demonstration of an AI-Based Satellite Attitude Controller}

\author{\uppercase{Kirill Djebko}\authorrefmark{1,2},
\uppercase{Tom Baumann}\authorrefmark{2},
\uppercase{Erik Dilger}\authorrefmark{2},
\uppercase{Frank Puppe}\authorrefmark{1}, and
\uppercase{Sergio Montenegro}\authorrefmark{2}}

\address[1]{Center for Artificial Intelligence and Data Science, Julius-Maximilians-Universität Würzburg, 97074 Würzburg, Germany}
\address[2]{Chair of Computer Science VIII - Aerospace Information Technology, Julius-Maximilians-Universität Würzburg, 97074 Würzburg, Germany}

\tfootnote{This work was conducted as part of the ``In-Orbit-Demonstrator Lernende Lageregelung (LeLaR)'' project (grant number/FKZ: 50RA2403), funded by the German Federal Ministry for Economic Affairs and Energy (Bundesministerium für Wirtschaft und Energie, BMWE) on the basis of a decision by the German Bundestag. Data Availability: Maneuver telemetry datasets are available at \underline{https://github.com/kdjebko/lelar-in-orbit-data}.}

\markboth
{Djebko \headeretal: LeLaR: The First In-Orbit Demonstration of an AI-Based Satellite Attitude Controller}
{Djebko \headeretal: LeLaR: The First In-Orbit Demonstration of an AI-Based Satellite Attitude Controller}

\corresp{Corresponding author: Kirill Djebko (e-mail: kirill.djebko@uni-wuerzburg.de).}

\begin{abstract}
Attitude control is essential for many satellite missions. Classical controllers, however, are time-consuming to design and sensitive to model uncertainties and variations in operational boundary conditions. Deep reinforcement learning offers a promising alternative by learning adaptive control strategies through autonomous interaction with a simulation environment. Overcoming the simulation-to-reality gap, which involves deploying an agent trained in simulation onto the real physical satellite, remains a significant challenge. In this work, we present the first successful in-orbit demonstration of an artificial intelligence-based attitude controller for inertial pointing maneuvers overcoming the simulation-to-reality gap. The controller was trained entirely in simulation and deployed to the InnoCube three-unit nanosatellite, which was developed by the Julius-Maximilians-Universität Würzburg in cooperation with the Technische Universität Berlin, and launched in January 2025. We present the artificial intelligence agent design, the training methodology, and a characterization of the discrepancies between the simulation environment and the in-orbit behavior of the satellite. To provide a reference, we evaluate the in-orbit performance of the artificial intelligence-based controller against both simulation results and the in-orbit performance of InnoCube’s default proportional-derivative controller under identical mission conditions. Steady-state metrics confirm the robust performance of the artificial intelligence-based controller during repeated in-orbit maneuvers.
\end{abstract}


\titlepgskip=-21pt

\maketitle
\begin{textblock}{130}(40,259)
\footnotesize\itshape\centering
\textcopyright~2026 The Authors. This article has been accepted for publication in IEEE Access (DOI: \href{https://doi.org/10.1109/ACCESS.2026.3678816}{10.1109/ACCESS.2026.3678816}).
This is the author's version which has not been fully edited and content may change prior to final publication. The final published version is available at \url{https://ieeexplore.ieee.org/document/11457578}.
Licensed under a Creative Commons Attribution 4.0 License. For more information, see \url{https://creativecommons.org/licenses/by/4.0/}.
\end{textblock}

\section{Introduction and Background}\label{sec:introduction}

In recent years, Machine Learning (ML) and Artificial Intelligence (AI)-based methods have increasingly entered engineering disciplines, particularly robotics, with notable progress in quadrupedal~\cite{pmlr-v164-rudin22a} and humanoid robots, such as Agile~Justin, developed at the German Aerospace Center (DLR)~\cite{R_stel_2025}. This influence is now expanding into the aerospace sector, where AI-driven methods are being explored for highly nonlinear applications, including robust attitude control for launch vehicles~\cite{aerospace12030203, Xue2023}, combustion chamber regulation~\cite{Hoerger2025}, UAV control~\cite{aerospace12080684, Mohiuddin2025}, and satellite attitude control~\cite{doi:10.2514/6.2025-1145, upm90311, retagne2024adaptive}. The motivation stems from the increasing autonomy and complexity required by modern spacecraft, which demand control strategies capable of adapting to dynamic environments while minimizing human intervention. Deep Reinforcement Learning (DRL), a variant of classical Reinforcement Learning~\cite{sutton1998reinforcement}, has shown considerable promise in this context. For example, for launch vehicles, DRL-based controllers using the Soft Actor–Critic (SAC)~\cite{haarnoja2018soft} algorithm have been applied in simulation to achieve load-relief attitude control under aerodynamic disturbances~\cite{aerospace12030203}. Similarly, DRL-based adaptive controllers for quad-tiltrotor UAVs improve disturbance rejection and tracking performance compared to classical methods~\cite{aerospace12080684} in simulation. For satellites, DRL has been used in simulation to maintain operational capability under unknown ADCS failures, where a Deep Deterministic Policy Gradient (DDPG)~\cite{lillicrap2015continuous}-based controller reorients solar panels toward the Sun without requiring explicit fault models~\cite{doi:10.2514/6.2025-1145}. Other work demonstrates the feasibility of real-time DRL-based magnetic attitude control on embedded hardware on ground~\cite{upm90311}. In addition, Proximal Policy Optimization (PPO)~\cite{schulman2017proximal}-based controllers using stacked observations, similar to our observation space design from~\cite{djebko2023learning}, achieve robust attitude control across varying satellite masses in simulation, outperforming classical PID control in settling time and stability~\cite{retagne2024adaptive}.

The Hybrid Online Policy Adaptation Strategy (HOPAS) project attempted an early in-orbit demonstration of a Reinforcement Learning-based attitude controller on the ESA OPS-SAT CubeSat, with the AI agent designed to modify PID control signals as a residual command, but the full AI closed-loop control was ultimately restricted to simulation and ground-testing~\cite{CarrilloBarrenechea2023, Airbus2022HOPAS}.
These developments illustrate the growing interest in DRL and AI-based control methods for aerospace systems characterized by nonlinear dynamics and uncertainties.

To date AI-based attitude control has been restricted to simulation environments or laboratory testbeds, with no successful in-orbit demonstrations. This work represents the first successful test of an AI-based attitude controller in orbit.

Building on our previous works~\cite{djebko2023learning, gerlich2023its, djebko2025}, we developed an AI-based attitude controller for the InnoCube 3U nanosatellite~\cite{aerospace8050127, montenegro2022innocube, baumann2025innocube}, developed by the Julius-Maximilians-Universität Würzburg and the Technische Universität Berlin, which successfully launched on January 14, 2025. The AI agent autonomously interacts with a simulated environment to learn a control strategy that maximizes a given reward function. The agent was first trained on a basic control task to obtain a base-agent, which was subsequently post-trained to produce the final flight-agent. The controller was trained entirely on ground, uploaded to InnoCube, and tested in orbit. Since this work was done as part of the ``In-Orbit Demonstrator Learning Attitude Control'' (German: In-Orbit-Demonstrator Lernende Lageregelung; LeLaR) project, we also refer to the AI agent as LeLaR.

A disadvantage of DRL lies in the tendency of neural networks to achieve good results in interpolation tasks but failure in extrapolation problems~\cite{tsimenidis2020limitationsdeepneuralnetworks}. Transferring an AI agent trained in simulation to the real target system, known as Sim2Real transfer, poses a significant challenge. Achieving zero-shot transfer, where the agent operates successfully upon deployment without further training, is critical, as maintaining high-fidelity models for every environmental contingency or performing iterative fine-tuning post-launch is often infeasible. For practical applications, it is therefore necessary to vary the boundary conditions during training, e.g., via domain randomization, to avoid so-called Out-of-Distribution errors. This includes, in particular, the moments of inertia, which were successfully addressed in our previous work~\cite{djebko2023learning, gerlich2023its} and are utilized here.

While originally a pure reaction wheel (RW) control was planned, an unexpected limitation regarding the wheel speeds emerged prior to the launch of InnoCube, which led to the additional consideration of magnetorquers (MT) in the controller design. Operation of the RWs in the range of [-350, 350] rpm should be avoided, as this can lead to inaccurate wheel speed estimation and control. As a precautionary measure, we have therefore split the agent network into a RW subnetwork and a MT subnetwork and integrated momentum management using the latter. During initial tests in orbit, several operational constraints limited our planned approach. In the early experiment phase, command list execution was not available, limiting tests to live teleoperation during passes. These passes, with a maximum duration of approximately 12 minutes, were too short to conduct complete momentum management tests and prompted us to consider the RW subnetwork in isolation. During these tests, however, it became apparent that the RW subnetwork's performance alone was sufficient and also fit the later safety and power constraint of a maximum experiment duration of 15 minutes after unsupervised command execution became available. Subsequently we used the RW subnetwork from the original agent as base-agent and adapted it to the remaining operational requirement changes by finetuning through post-training as we lay out in Section~\ref{sec:flight_agent}.

This paper is structured as follows: Section~\ref{sec:methodology} discusses the methodology, including the satellite model, agent design, and Safety Cage. Section~\ref{sec:results} presents Sim2Real discrepancies and in-orbit experiments. Section~\ref{sec:conclusions} concludes with a discussion and provides an outlook on future work.

\section{Methodology}\label{sec:methodology}

The following sections outline the components required to apply DRL to the attitude control task. This involves developing a simulation model of the target system, defining the observation space as the AI agent’s inputs and the action space as its outputs, and specifying an appropriate optimization target. This optimization target, the so-called reward function, together with the observation space, form the core components of the DRL procedure. For training we follow an iterative approach by first training a base-agent that learns to solve the basic attitude control problem and which is then post-trained to handle more complicated boundary conditions, augmented by noise and disturbances.

As base-agent, we employ the post-trained agent from our prior work~\cite{Djebko2025_SubnetworkPolicy}, using a split policy network, consisting of an RW and MT subnetwork. It was originally designed to avoid prolonged RW operation in the interval of $[-350,350]$~\text{rpm} to avoid the inaccuracies stemming from the RW rate estimator when operated in this range. Initial tests in-orbit however revealed that the RW subnetwork alone performed sufficiently well, as we will discuss in Section~\ref{sec:results}. Along with changed safety requirements introduced after deployment but prior to the first test, this observation prompted us to use the RW subnetwork of the agent from~\cite{Djebko2025_SubnetworkPolicy} as base and to fine tune it through post-training for standalone use. We refer to this agent as LeLaR flight-agent or flight-agent in short.

Since~\cite{Djebko2025_SubnetworkPolicy} is available only in German, we adapt and summarize the essential elements required to understand the current work and expand on important aspects. This includes the base-agent design and the sensor calibration methodology adapted from our prior work, covering both the physical on-board sensors and the simulation environment used for training of all agents. It is to be noted that sensor calibration is not a requirement introduced by the DRL procedure but should generally be performed for any satellite mission. It is included here for completeness, as calibration affects both the satellite's sensors and the quality of the simulation. We then highlight the modifications to the model and reward function, as well as the updated operational requirements and boundary conditions, applied to obtain the flight-agent. Finally, the Safety Cage architecture is described, which monitors telemetry against predefined safety limits and triggers a fallback to a safe idle state in case of anomalous behavior during AI-controlled maneuvers by immediately turning off the AI-based controller.

\subsection{InnoCube Mission and ADCS Architecture}\label{sec:innocube}

InnoCube is a 3U CubeSat mission designed as a technology demonstrator for several innovative technologies and has successfully been launched on January 14, 2025 into a 520~km sun-synchronous orbit. It has a mass of 4.2 kg and carries two main experiments: ``Skip The Harness'' (SKITH), a fully wireless data bus for intra-satellite communication and ``Fibre-Reinforced Spacecraft Wall for Storing Energy'' (Wall\#E), an experimental carbon fiber reinforced plastic solid-state battery. Secondary payloads include amateur radio payloads and ``Experiment for Precise Orbit Determination'' (EPISODE), a software defined GNSS receiver. An overview of the spacecraft’s subsystems and structure is provided in Fig.~\ref{fig:innocube_overview}. The InnoCube project is funded by the German Federal Ministry for Economic Affairs and Energy (BMWE) under grant number (FKZ) 50RU2000.

\Figure[t!](topskip=0pt, botskip=0pt, midskip=0pt)[width=0.99\linewidth]{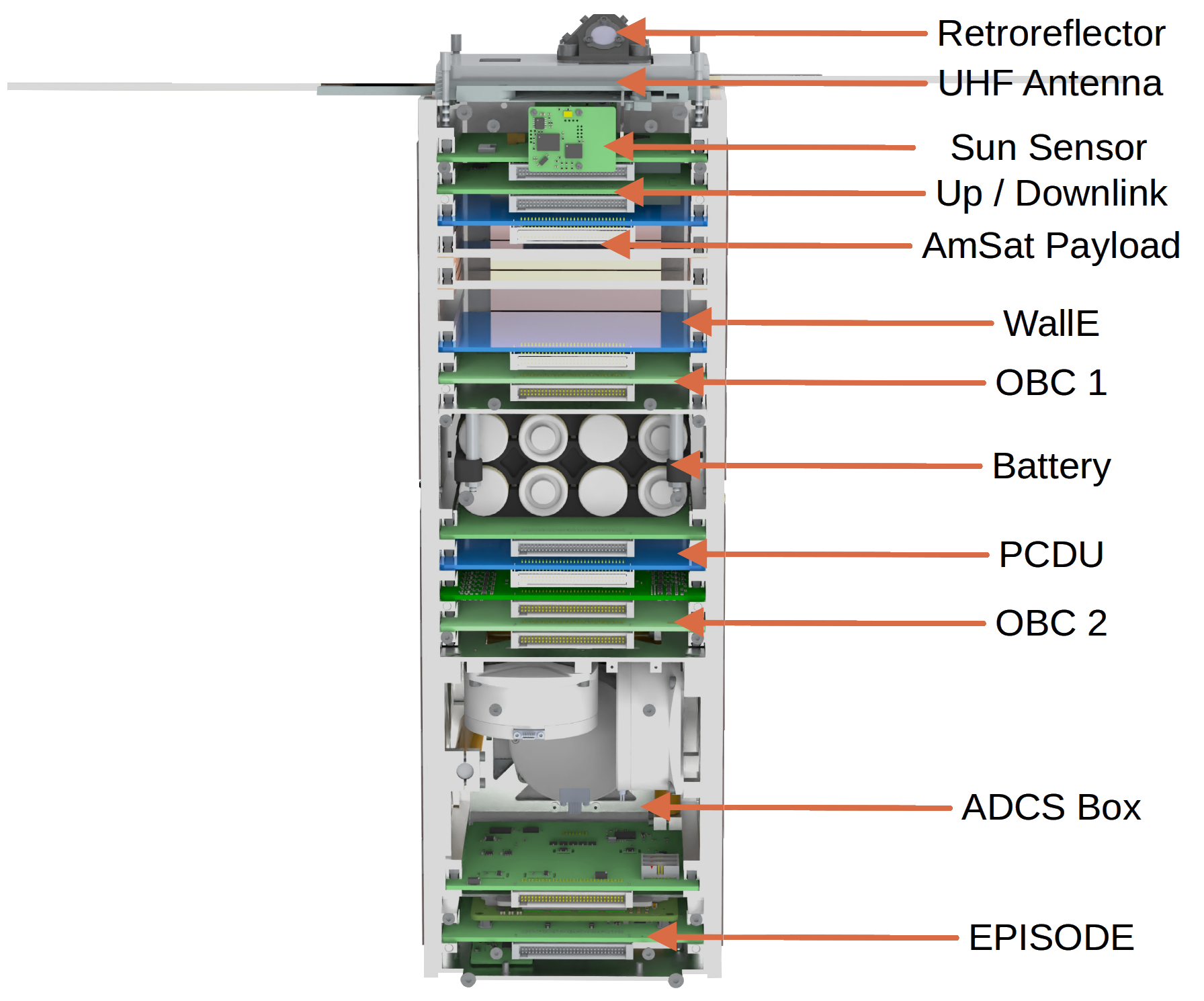}
{\textbf{Overview of the components of the InnoCube satellite.} \label{fig:innocube_overview}}

\subsubsection{InnoCube ADCS}\label{sec:innocube_adcs}

The Attitude Determination and Control System (ADCS) is housed within a dedicated sub-assembly (ADCS-Box). It comprises the cold-redundant ADCS mainboard, including gyro and magnetometer sensors, three ferrite-core magnetorquers with dual redundancy along all axes, and three reaction wheels. Sun sensors mounted on the exterior of the satellite provide external sensing and are connected wirelessly to the ADCS mainboard.

The ADCS for InnoCube was primarily developed to meet the requirements of the EPISODE payload, which imposed relatively modest attitude performance constraints. Two operational modes were considered: GNSS Pointing and Laser-Ranging Tracking. GNSS Pointing requires the GNSS Antennas to be pointed away from Earth, providing unobstructed view for the antennas to GNSS satellites. This resulted in a relatively low pointing requirement of around $\pm90^\circ$ of anti-nadir pointing. Laser-Ranging Tracking requires the retroreflector to be pointed toward a ground station with a pointing accuracy of $\pm17^\circ$ during a ground station pass. The LeLaR project began after the design of the InnoCube satellite was completed. Consequently, design requirements for LeLaR could only be based on the existing InnoCube system.

The ADCS mainboard is based on an EFR32FG12 microcontroller~\cite{efr32fg12} and is extended with the electrical components required for ADCS operation. These include a STMicroelectronics ASM330LHH micro-electromechanical systems (MEMS) gyroscope~\cite{asm330gyro} for body-rate measurements, a three-axis PNI RM3100 magneto-inductive magnetometer~\cite{rm3100mag} for magnetic field sensing, controllers and interfaces for the MTs, and interfaces to the RWs. The external sun sensors are not used in the LeLaR experiments of this work, as relative attitude is determined solely via quaternion integration of the measured gyroscope body rates.

Since InnoCube lacks highly accurate absolute determination sensors (e.g., star-trackers) and the accuracy of the sensor fusion utilizing magnetometer and sun-sensor measurements currently does not produce accurate continuous determination-solutions, the in-orbit data rely on gyro measurements only. This has the advantage, that the evaluation of control accuracy will only be affected by the accuracy of the gyro integration. Since the analyses of this work focus on relative control performance only, absolute attitude knowledge is not strictly required for this evaluation. As described in Section~\ref{sec:results}, prior to executing maneuvers, the initial quaternion for a maneuver is set to a specific quaternion, which indicates a deviation from the goal attitude, being the identity quaternion $(1, 0, 0, 0)$ in all cases. The maneuver is then executed in the respective inertial frame as set by the initial quaternion, which has no absolute orientation information. The relative error trajectory is invariant to the choice of the initial quaternion.

Attitude control on InnoCube is achieved using the three RWs and MTs, arranged orthogonally along the satellite principal axes. The RWs provide the main attitude control, while the MTs are used for wheel momentum management and detumbling during and after maneuvers. The RWs offer a maximum torque of 2~\text{mN}$\cdot$\text{m} and a maximum angular momentum of 30~\text{mN}$\cdot$\text{m}$\cdot$\text{s} per axis. The in-orbit RW behavior is discussed in Section \ref{sec:rw_behaviour}. The MTs support a maximum commandable dipole moment of approximately $\pm$0.35~\text{A}$\cdot$\text{m}$^2$ along each axis. A summary of the ADCS specifications is presented in Table \ref{tab:adcs_specs}.

\begin{table}[t!]
\caption{\textbf{InnoCube ADCS hardware specifications.}}
\label{tab:adcs_specs}
\setlength{\tabcolsep}{3pt}
\begin{tabular}{p{75pt} p{152pt}}
\hline
Component & Specification \\
\hline
Computing Unit & SL EFR32FG12 (40~MHz, 256~kB RAM) \\
Gyroscope & ST ASM330LHH (1.41 $\times$ 10$^{-4}$~\text{rad/s} noise) \\
Magnetometer & PNI RM3100 (5.1 $\times$ 10$^{-3}$~$\mu$\text{T} noise) \\
Reaction Wheels & $3\times$ orthogonal (max 2~\text{mN}$\cdot$\text{m}, 30~\text{mN}$\cdot$\text{m}$\cdot$\text{s}) \\
Magnetorquers & $3\times$ Ferrite-Core (approx. $\pm$0.35~\text{A}$\cdot$\text{m}$^2$) \\
Determination & Relative rates integration \\
Control Freq. & 1~Hz control loop \\
\hline
\end{tabular}
\end{table}

\Figure[t!](topskip=0pt, botskip=0pt, midskip=0pt)[width=0.99\linewidth]{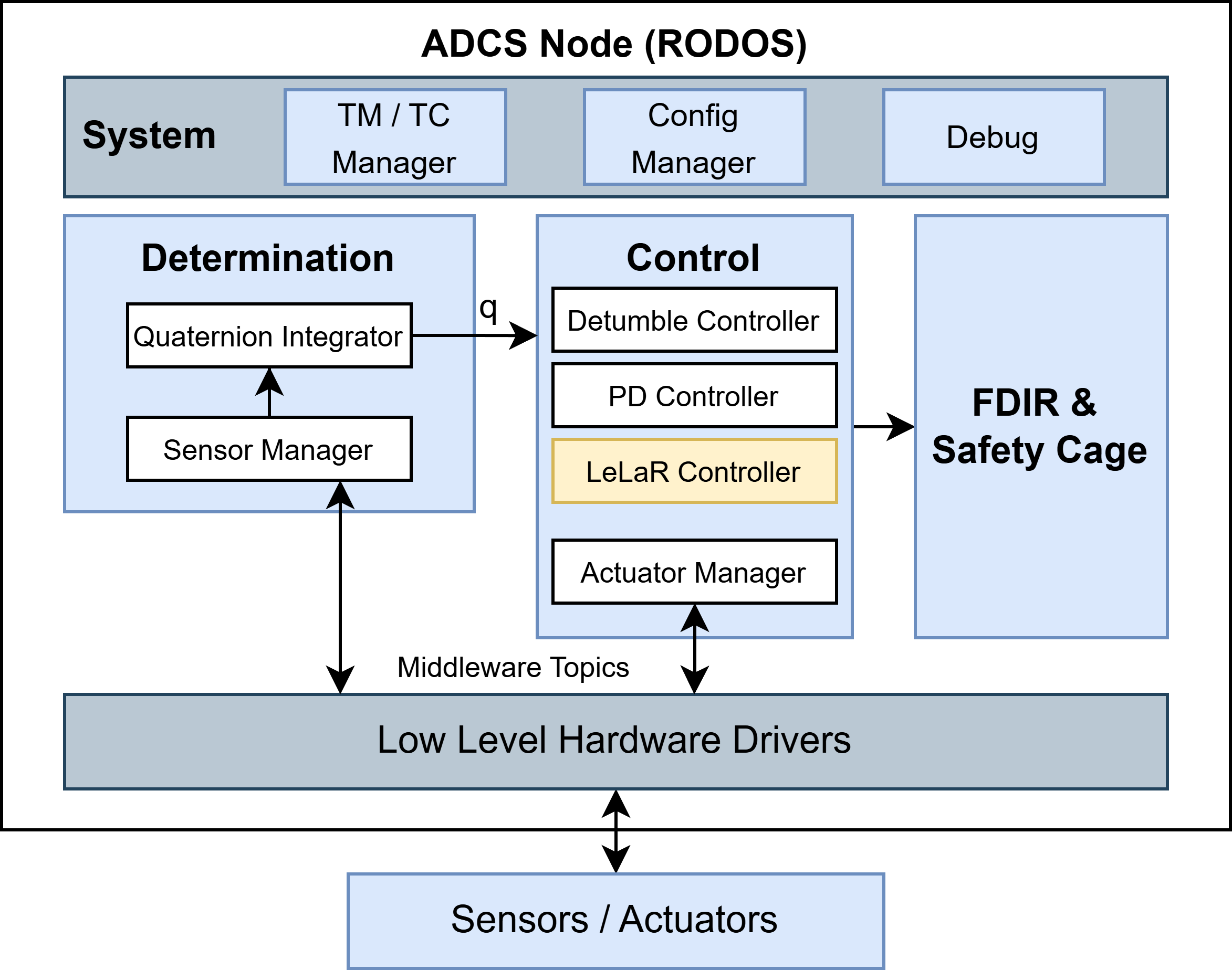}
{\textbf{Schematic depiction of the ADCS software modules.}\label{fig:adcs_software}}

The InnoCube ADCS software is constructed in a modular manner. Fig.~\ref{fig:adcs_software} depicts the different software components. The software runs on the on-board operating system RODOS. The determination module collects data from the low-level hardware drivers, which connect to the sensors. Since the LeLaR experiments currently focus only on control performance and not on overall system performance (which includes determination performance), the relative attitude is estimated only using gyro rate integration in the quaternion integrator module. The determination module outputs a calculated attitude quaternion along with the current satellite body rates. This information, together with all current sensor values, is forwarded into the controller module, which can host several control algorithms. Along with a commanded attitude, the control module calculates a control command, which is then forwarded to the satellite's actuators. Fault detection isolation and recovery (FDIR) as well as the Safety Cage architecture (see Section~\ref{sec:safetycage}) run in parallel in a separate module. The whole ADCS loop runs with a frequency of 1 Hz.

\subsubsection{InnoCube default PD Controller}\label{sec:innocube_pd}
For inertial pointing maneuvers, a classical proportional-derivative (PD) controller based on the following control law has been implemented, serving as the main controller for the original InnoCube mission.

With $\mathbf{q}$ mapping the inertial to the satellite-fixed body frame and a constant target $\mathbf{q}_\mathrm{t}$, the attitude error is:
\begin{equation}
\label{eq:error_q}
\delta \mathbf{q}
=
\begin{bmatrix}
\delta q_\mathrm{s} \\
\delta \mathbf{q}_\mathrm{v}
\end{bmatrix}
=
 \mathbf{q}_\mathrm{t}^{-1} \odot \mathbf{q},
\end{equation}

where $\delta q_\mathrm{s} = \delta q_0$ is the scalar part of the attitude error quaternion, $\delta \mathbf{q}_\mathrm{v} = [\delta q_1,~ \delta q_2,~ \delta q_3]^T$ denotes its vector part and $\odot$ is the  Hamilton quaternion product. The commanded body torque $\boldsymbol{\tau}_{\mathrm{B}}$ is defined using the following control law (see also~\cite{markley_fundamentals_2014}):
\begin{equation}
\boldsymbol{\tau}_{\mathrm{B}} = -\mathbf{k}_p \mathrm{sign}(\delta q_\mathrm{s}) \delta \mathbf{q}_\mathrm{v}
- \mathbf{k}_d (1 + \delta \mathbf{q}_\mathrm{v}^T \delta \mathbf{q}_\mathrm{v}) \boldsymbol{\omega},
\label{eq:pd}
\end{equation}
where $\boldsymbol{\omega}$ is the rotational rate vector in the satellite body frame in radians per second. The proportional gain is defined as the vector $\mathbf{k}_\mathrm{p}$, where 
\begin{equation}
\mathbf{k}_\mathrm{p} =
\begin{bmatrix}
\widetilde{k_\mathrm{p}} \\
\widetilde{k_\mathrm{p}} \\
\widetilde{k_\mathrm{p}} ~ z_{\mathrm{axis\_scale}}
\end{bmatrix}
\end{equation}

and similarly the derivative gain $\mathbf{k}_\mathrm{d}$ is defined as:

\begin{equation}
\mathbf{k}_\mathrm{d} =
\begin{bmatrix}
\widetilde{k_\mathrm{d}} \\
\widetilde{k_\mathrm{d}} \\
\widetilde{k_\mathrm{d}} ~ z_{\mathrm{axis\_scale}}
\end{bmatrix}
\end{equation}

The $\widetilde{k_\mathrm{p}}$ and $\widetilde{k_\mathrm{d}}$ parameters are positive scalar gains of $\widetilde{k_\mathrm{p}} = 3.99 \times10^{-3}$ and $\widetilde{k_\mathrm{d}} = 1.49 \times10^{-2}$. The PD controller gain values have been manually derived and tuned using a similar simulation setup to that used for LeLaR. No particular emphasis on high controller performance had been placed on the controller, given the requirements as mentioned in Section~\ref{sec:innocube_adcs}. 

The Z-axis scale is approximated using the ratio of the principal inertial moments of the Z-axis with respect to the X-axis of the satellite. This factor approximates to around 0.2, which has been used to scale the Z-axis torque correspondingly. The PD controller had already been in use prior to the LeLaR experiments on-board of InnoCube and was developed independently from LeLaR.

\subsection{Simulation Model and System Parameters}\label{sec:simulationmodel}
A simulation model of the InnoCube satellite was built using component specifications, CAD-based inertia estimates, and experimental subsystem data, as initially detailed in~\cite{Djebko2025_SubnetworkPolicy}. The simulated orbit is based on InnoCube TLE data from CelesTrak~\cite{celestrak2025innocube}, with variations applied during training. Table~\ref{tab:system_params} lists the key model parameters relevant to the dynamics of the model and summarizes nominal values and offsets used for domain randomization during agent training. Observations and actions are computed at 1~Hz and system dynamics are simulated at 10~Hz.

\begin{table}[t!]
\caption{\textbf{Key Satellite and Orbital Parameters With Training Variations ($U$ Denotes a Uniform Distribution).}}
\label{tab:system_params}
\setlength{\tabcolsep}{3pt}
\begin{tabular}{p{55pt} p{110pt} p{60pt}}
\hline
Parameter & Value & Variation (offset) \\
\hline
\multicolumn{3}{l}{\textit{Dynamics}} \\
Sat. Inertia & $[4.28, 4.22, 0.985] \times 10^{-2}$~\text{kg}$\cdot$\text{m}$^2$ & $\pm 15\%$ (per axis) \\
RW Max Speed & $1.64 \times 10^4$~\text{rpm} & -- \\
RW Inertia & $5.68 \times 10^{-5}$~\text{kg}$\cdot$\text{m}$^2$ & -- \\
RW min Torque & $10^{-5}$~\text{N}$\cdot$\text{m} & -- \\
RW max Torque & $2 \times 10^{-3}$~\text{N}$\cdot$\text{m} & -- \\
\hline
\multicolumn{3}{l}{\textit{Orbital}} \\
Perigee& $508$ km & $U(-5, 5)$ km \\
Apogee & $519$ km & $U(-5, 5)$ km \\
Eccentricity & $7.63 \times 10^{-4}$ & $U(-1, 3) \times 10^{-4}$ \\
Inclination & $97.43^\circ$ & $U(-0.03, 0.03)^\circ$ \\
RAAN & $U(0, 360)^\circ$ & -- \\
Arg. of Periapsis & $U(0, 360)^\circ$ & -- \\
True Anomaly & $U(0, 360)^\circ$ & -- \\
\hline
\end{tabular}
\end{table}

\subsection{Observation and Action Space}\label{sec:observationspace}

The observation space is defined as a 39-dimensional vector, normalized to $[-1,1]$. The action space of the base-agent (see Section~\ref{sec:base_agent}) has 6 elements: three RW torque commands and three MT dipole commands. The action space of the flight-agent (RW subnetwork; see Section~\ref{sec:flight_agent}) is subsequently the three RW torque commands. The agent output values are in the interval of $[-1,1]$. To obtain the effective action the agent outputs have to be scaled by the maximum RW torque or maximum dipole moment of the respective actuator. The reasoning behind this observation space design is that the agent is effectively given two consecutive system states as well as the last performed RW action, which allows it to calculate the dynamics of the system from the system state delta. Table~\ref{tab:observation_space} shows the different components of the AI agent's observation space. Note that if $\delta q_0 < 0$ then $\delta \mathbf{q}$ is multiplied by $-1$ prior to being input to the AI agent to address the unwinding problem.
\begin{table}[t!]
\caption{\textbf{Features of the Observation Space.}}
\label{tab:observation_space}
\setlength{\tabcolsep}{3pt} 
\begin{tabular}{p{45pt}p{145pt}p{35pt}}
\hline
Component & Description (Length) & Time \\
\hline
$\mathbf{err\_quat}$ & Attitude error quaternion (4) & $t_i, t_{i-1}$ \\
$\mathbf{err\_rr}$   & Satellite rate error (3) & $t_i, t_{i-1}$ \\
$\mathbf{a\_rw}$     & Last executed RW action (3) & $t_{i-1}$ \\
$\mathbf{rwr}$      & RW speeds (3) & $t_i, t_{i-1}$ \\
$\mathbf{mag\_b}$    & Magnetic field vector (3) & $t_i, t_{i-1}$ \\
$\mathbf{err\_rwr}$  & RW speed error (3) & $t_i, t_{i-1}$ \\
$\mathbf{crm}$      & Cross prod. $\mathbf{rwr} \times \mathbf{mag\_b}$ scaled by 0.5 (3) & $t_i$ \\
$\mathrm{bn}$       & Magnetic field norm (1) & $t_i$ \\
\hline
\end{tabular}
\end{table}

\subsection{Calibration}\label{sec:calibration}
To increase the fidelity of the simulation, available sensor data of the gyros and magnetometers were captured and used to augment the simulation and incorporated into the domain randomization during training.
With the availability of telemetry data after the satellite's launch, the moments of inertia could additionally be refined, and the residual magnetic dipole moments as well as the gyro bias recalibrated. While this is typically handled as part of regular mission operations using classical calibration methods, machine learning techniques were applied here to refine the initial calibration.

\subsubsection{Sensor Noise Model}\label{sec:sensornoisemodel}
To create the sensor noise model, measurement data from the gyroscopes (MEMS) and magnetometers (magneto-inductive) was recorded using the InnoCube EQM in a Thermal Vacuum Chamber (TVAC). The chamber was evacuated to $10^{-4}$~\text{mBar} to eliminate convection effects and cooled. Since the cooling units produce magnetic interference and vibration, data acquisition was performed only after reaching the target temperature with all pumps and cooling disabled. The EQM then warmed naturally back to approximately 30~$^\circ$\text{C} room temperature.

The dataset was cropped to match the in-orbit temperature range of -3.55~$^\circ$\text{C} to 15.6~$^\circ$\text{C}. In total, data from five experiments were evaluated and a sensor noise model was built. The noise model consists of a bias term $v_\mathrm{b}$, sampled once per episode (corresponding roughly to an attitude maneuver), and white noise $v_\mathrm{w}$, sampled every time step during simulation.

The raw data were divided into 60~s segments (artificial ``maneuvers''), assuming nearly constant bias within each segment. For each segment, the mean value was taken as the bias estimate. The standard deviation of the segment means forms $\sigma_\mathrm{b}$. Since temperatures near room temperature are overrepresented, segment means were temperature-weighted based on their histogram distribution to avoid bias toward those ranges. White noise statistics were obtained by subtracting each segment mean from the corresponding samples. The standard deviation of all residuals across all segments forms $\sigma_\mathrm{w}$. For all simulations, both $\sigma_\mathrm{b}$ and $\sigma_\mathrm{w}$ were additionally scaled by factor of 1.2 prior to sampling for domain randomization. The noise sampling model can in brief be described by \eqref{eq:vb}--\eqref{eq:v_total}:
\begin{equation}
    v_\mathrm{b} \sim \mathcal{N}(0, \sigma_\mathrm{b}^2)
    \label{eq:vb}
\end{equation}
\begin{equation}
    v_\mathrm{w} \sim \mathcal{N}(0, \sigma_\mathrm{w}^2)
    \label{eq:vn}
\end{equation}
\begin{equation}
    v_{\mathrm{eff}} = v_{\mathrm{sim}} + v_\mathrm{b} + v_\mathrm{w},
    \label{eq:v_total}
\end{equation}

where $v_{\mathrm{sim}}$ denotes the noise-free sensor value from the simulation and $v_{\mathrm{eff}}$ is the resulting effective noisy sensor value used as input to the AI agent during training. The resulting parameters, derived exclusively from TVAC data and without any temperature compensation applied, are given in Table~\ref{tab:noise_parameters}. These bias variations deliberately contain the complete uncompensated temperature drift over the full temperature cycle. This approximates the disturbance the AI agent is expected to encounter in orbit.

\begin{table}[t!]
\caption{\textbf{Sensor Noise Parameters (Including Domain-Randomization Scaling With Factor 1.2) Derived From TVAC Testing.}}
\label{tab:noise_parameters}
\setlength{\tabcolsep}{3pt} 
\begin{tabular}{p{110pt} p{35pt} p{35pt} p{35pt}}
\hline
Parameter & X-axis & Y-axis & Z-axis \\
\hline
\multicolumn{4}{l}{Gyroscopes ($10^{-2}$ $^\circ$\text{/s})} \\
Bias std $\sigma_\mathrm{b}$ & 2.01 & 1.42 & 9.77 \\
White-noise std $\sigma_{\mathrm{w}}$ & 1.38 & 1.12 & 1.06 \\
\hline
\multicolumn{4}{l}{Magnetometers ($10^{-2}\,\mu\text{T}$)} \\
Bias std $\sigma_\mathrm{b}$ & 5.48 & 8.48 & 7.58 \\
White-noise std $\sigma_{\mathrm{w}}$ & 0.865 & 0.907 & 1.13 \\
\hline
\end{tabular}
\end{table}

\subsubsection{Moments of Inertia Estimation}
The satellite's moments of inertia ($\mathbf{I}$) were determined through a combination of CAD modeling, telemetry-based estimation, and analytical verification. The final tensor, $\mathbf{I} = \text{diag}(0.0428, 0.0422, 0.00985)$~\text{kg}$\cdot$\text{m}$^2$, deviated by at most 2.5\% from initial CAD estimates. To ensure controller robustness, $\mathbf{I}$ was varied by $\pm$15\% per axis during domain randomization in the training phases of all agents. Note that for the InnoCube satellite, the principal axis frame very closely aligns with the satellite body frame given its symmetrical 3U CubeSat structure and even mass distribution. The off-diagonal elements as calculated from the CAD model can be considered negligible.

\subsubsection{Residual Dipole Moment Estimation}\label{sec:resdipole}
InnoCube exhibited unexpectedly high residual magnetic dipole moments ($\boldsymbol{\mu}$) during early in-orbit operations. To mitigate the resulting disturbance torques $\mathbf{T}_{\mu}$, we developed a system calibration framework leveraging automatic differentiation. The residual dipole and gyroscope bias ($\mathbf{b}$) were estimated by minimizing the residual of Euler’s rotational dynamics~\cite{BUSCH201573}:
\begin{equation}
    \mathbf{T}_{\mathrm{ext}} = \mathbf{I}\dot{\boldsymbol{\omega}} + \boldsymbol{\omega} \times (\mathbf{I}\boldsymbol{\omega})
\end{equation}
\begin{equation}
    \mathbf{T}_{\mu} = \boldsymbol{\mu} \times \mathbf{B}
\end{equation}

The optimization assumes that the dominant external torque is due to the residual magnetic dipole:
\begin{equation}
    \mathbf{T}_{\mathrm{ext}} \overset{!}{=} \mathbf{T}_{\mu}
\end{equation}

Where $\mathbf{T}_{\mathrm{ext}}$ is the external torque acting upon the satellite and $\mathbf{T}_{\mu}$ is assumed to be the main contributing factor.
The optimization objective was formulated as:
\begin{equation}
    \min_{\boldsymbol{\mu}, \mathbf{b}} \sum \| \mathbf{I}\dot{\boldsymbol{\omega}} + \boldsymbol{\omega}_{\mathrm{corr}} \times (\mathbf{I}\boldsymbol{\omega}_{\mathrm{corr}}) - (\boldsymbol{\mu} \times \mathbf{B}) \|^2,
\end{equation}
where $\boldsymbol{\omega}_{\mathrm{meas}}$ is the raw angular velocity measured by the onboard gyroscope, $\mathbf{b}$ is the gyroscope bias vector to be estimated, and $\boldsymbol{\omega}_{\mathrm{corr}} = \boldsymbol{\omega}_{\mathrm{meas}} + \mathbf{b}$ is the bias-corrected angular velocity. Further $\mathbf{I}$ is the inertia tensor and $\mathbf{B}$ is the measured magnetic field in body frame components. We utilized the AdamW optimizer and PyTorch's autograd engine~\cite{paszke2019pytorchimperativestylehighperformance} to train the parameters, yielding $\boldsymbol{\mu} = [-0.459, -0.024, 0.069]$~\text{A}$\cdot$\text{m}$^2$ and $\mathbf{b} = [-0.028, 0.761, -0.032]~^\circ$\text{/s}. 

Further validation was performed by activating MTs sequentially, obtaining the corresponding telemetry data and re-running the PyTorch script on this data. The resulting magnetic-field changes corresponded to dipole moments of 0.31--0.39~\text{A}$\cdot$\text{m}$^2$, which was in accordance with the expected theoretical values, thus confirming the MT characteristics and the correctness of the learned $\boldsymbol{\mu}$. The learned gyro bias was applied and replaced previous calibration for all gyro data processing of the InnoCube mission. For the training of the base-agent, the active residual dipole moment compensation was simulated with a random error of up to $\pm$10\% for each axis independently and up to $\pm$25\% for the flight-agent.

Following the application of the dipole moment compensation from one set of the MTs, in-orbit checks showed a significant reduction of the residual dipole moment, reducing the buildup of angular momentum during maneuvers. To fully eliminate the residual in the X-axis, the second set of MTs has to be used in the future, since the required dipole moment exceeds the available dipole moment per single axis set. Although initially intended, this is currently not implemented for safety reasons, as this would require running both ADCS nodes simultaneously.

\subsection{Base-Agent Design}\label{sec:base_agent}
We used our AI agent from~\cite{Djebko2025_SubnetworkPolicy} as a base, and derived the flight-agent through post-training, which is conceptually related to transfer learning. While the base-agent was initially trained to solve a different variation of the attitude control problem, we employed post-training with modified boundary conditions and a revised optimization target to obtain the flight-agent. This approach leverages the well-initialized policy that already captured the fundamental dynamics of the system, avoiding redundant re-learning from scratch. As outlined in Section~\ref{sec:introduction} the flight-agent is based off of the RW subnetwork of the base-agent. In the following section, the base-agent training is described. The modifications performed to obtain the flight-agent are explained in Section~\ref{sec:flight_agent}.

\subsubsection{Subnetwork Policy}\label{sec:twin-network-policy}
The base-agent employed a split policy network consisting of one RW subnetwork and one MT subnetwork, producing a six-element action vector by concatenating the three RW actions and the three MT actions. Each subnetwork can be in one of three states: DISABLED, FROZEN, or ACTIVE. DISABLED outputs zeros, does not contribute to training, and has non-trainable weights; FROZEN outputs effective actions but is not trainable; ACTIVE outputs effective actions and fully participates in training.

\subsubsection{Base-Agent Reward Functions}\label{sec:rewardfunction}

The base-agent was trained with a PPO derivative, SkipPPO~\cite{Djebko2025_SubnetworkPolicy}, utilizing the separate RW and MT subnetworks. The RWs learn to quickly attain the goal attitudes, whereas the MTs learn slow momentum management to bring the RW speeds to $\pm$500~\text{rpm} (corresponding to $\pm$350~\text{rpm} and a safety buffer). 
The RW reward function of the base-agent rewards quick reduction of attitude error, reduction of angular rates, and stable convergence. The MT reward encourages minimizing RW speed deviations. A combined reward was used during post-training of the base-agent to balance both effects. The RW reward function is shown in \eqref{eq:base_agent_rw_reward_1}--\eqref{eq:base_agent_rw_reward_6}.

\begin{equation}
\mathrm{err\_att}_{t_i} = 1 - | \delta q_{0,t_i} |
\label{eq:base_agent_rw_reward_1}
\end{equation}
\begin{equation}
\Delta \theta_{t_i} = \mathrm{err\_att}_{t_i} - \mathrm{err\_att}_{t_{i-1}}
\label{eq:base_agent_rw_reward_2}
\end{equation}
\begin{equation}
\mathrm{p\_norm}_{t_i} = \| \mathbf{err\_rr}_{t_i} \|
\label{eq:base_agent_rw_reward_3}
\end{equation}
\begin{equation}
\mathrm{p\_att}_{t_i} = 0.1 \cdot \mathrm{err\_att}_{t_i}
\label{eq:base_agent_rw_reward_4}
\end{equation}
\begin{equation}
r_{t_i} =
\begin{cases}
1 + \frac{1}{\mathrm{p\_norm}_{t_i}+0.1}, & \mathrm{err\_att}_{t_i} < 3.8\cdot10^{-5} \\
e^{-\mathrm{err\_att}_{t_i}/0.14}, & \mathrm{err\_att}_{t_i} < \mathrm{err\_att}_{t_{i-1}} \\
0.1~ e^{-\Delta\theta_{t_i}/0.14}-1, & \Delta\theta_{t_i} < \Delta\theta_{t_{i-1}} \\
e^{-\mathrm{err\_att}_{t_i}/0.14} - 2, & \text{else}
\end{cases}
\label{eq:base_agent_rw_reward_5}
\end{equation}
\begin{equation}
\mathrm{reward\_rw}_{t_i} = r_{t_i} - \mathrm{p\_norm}_{t_i} - \mathrm{p\_att}_{t_i}
\label{eq:base_agent_rw_reward_6}
\end{equation}

$\delta q_0$ is the scalar component of the attitude error quaternion as defined in~\eqref{eq:error_q}, mapped to its positive scalar representation as described in Section~\ref{sec:observationspace}, ensuring $\delta q_{0,t_i} = |\delta q_{0,t_i}|$. The absolute value in~\eqref{eq:base_agent_rw_reward_1} is used to ease the understanding of this implicit relation. $\mathrm{p\_norm}$ ($p$ stands for penalty) penalizes satellite body rotation rates, and $\mathrm{p\_att}$ penalizes residual attitude error. The main reward term $r_{t_i}$ gives strong reward for attaining the target attitude (within $\approx1^\circ$), moderate reward for approaching it, slight penalties for overshooting correction and stronger penalties for diverging. The MT reward function is shown in \eqref{eq:base_agent_mt_reward_1}--\eqref{eq:base_agent_mt_reward_3}.
\begin{equation}
\mathrm{err\_deviation}_{t_i} = \sum \mathbf{err\_rwr}_{t_i} \cdot 16384.0
\label{eq:base_agent_mt_reward_1}
\end{equation}
\begin{equation}
\mathrm{p\_action}_{t_i} = \frac{\sum |\mathbf{a\_mt}_{t_i}|}{0.6 \cdot 50}
\label{eq:base_agent_mt_reward_2}
\end{equation}
\begin{equation}
\mathrm{reward\_mt}_{t_i} = \frac{1 - \mathrm{p\_action}_{t_i}}{\sqrt{\mathrm{err\_deviation}_{t_i} + 1}}
\label{eq:base_agent_mt_reward_3}
\end{equation}
Here, $\mathbf{a\_mt}_{t_i}$ are the current agent MT actions. The reward decreases with RW speed deviations and with MT action magnitude. It favors minimizing RW momentum with low dipole effort. For the base-agent post-training the reward function from \eqref{eq:base_agent_combined_reward} was implemented, scaling the RW reward to match the MT reward magnitude in order to avoid bias.
\begin{equation}
\mathrm{reward\_c}_{t_i} = \frac{\mathrm{reward\_rw}_{t_i}}{11} + \mathrm{reward\_mt}_{t_i}
\label{eq:base_agent_combined_reward}
\end{equation}

For training, the Stable-Baselines3~\cite{JMLR:v22:20-1364} library was employed and subsequently modified. The Basilisk Astrodynamics Simulation Framework~\cite{kenneally2020basilisk} was used as the simulator. To avoid oscillating control signals, generalized State Dependent Exploration (gSDE)~\cite{raffin2022smooth} was employed. The networks were initialized using orthogonal weight initialization, with scaling factors of 0.01 for the RW action network, 10.0 for the MT action network, and 1.0 for all remaining networks. The RW subnetwork consisted of three hidden layers of 64 neurons each for the policy and two hidden layers of 64 neurons each for the value network. The MT subnetwork used four hidden layers of 64 neurons each for both the policy and the value networks. All networks employed the SiLU activation function. Table~\ref{tab:rw_mt_hyperparams} shows the hyperparameters used. For a more detailed description of PPO and its hyperparameters we refer the reader to~\cite{schulman2017proximal, JMLR:v22:20-1364, raffin2022smooth}. Sections~\ref{sec:twin-network-policy}--\ref{sec:finetuning} describe the initial training and post-training process of the base-agent.

\begin{table}[t!]
\caption{\textbf{Hyperparameters for Base-Agent Training/Post-Training.}}
\setlength{\tabcolsep}{3pt} 
\label{tab:rw_mt_hyperparams}
\begin{tabular}{p{95pt} p{70pt} p{60pt}}
\hline
Hyperparameter & RW Network & MT Network \\ \hline
Total Time Steps & $5 \times 10^8$ & $5 \times 10^8$ \\
Rollout Steps ($n$) & $2 \times 10^3$ / $4 \times 10^4$ & $4 \times 10^4$ \\
$\gamma$ (Discount Factor) & 0.95 & 0.97 / 0.95 \\
Batch Size & 64 & 64 \\
Clip Range & 0.2 & 0.2 \\
Target KL$^{\mathrm{a}}$ & 0.2 & 0.2 \\
(g)SDE Sampling Freq. & 1 per rollout & 1 per rollout \\
Log STD Init.$^{\mathrm{b}}$ & -2.0 & 0.0 / -2.0 \\
Learning Rate & $1 \times 10^{-4}$ / $1 \times 10^{-5}$ & $5 \times 10^{-5}$ \\
Episode Length & 50 / 5000 & 5000 \\
$\mathrm{num\_skip}$ & 0 & 9 / 0 \\
Activation Function & SiLU & SiLU \\
Weight Initialization & Orthogonal & Orthogonal \\
\hline
\multicolumn{3}{p{215pt}}{\scriptsize $^{\mathrm{a}}$Target Kullback-Leibler Divergence for early stopping.} \\
\multicolumn{3}{p{215pt}}{\scriptsize $^{\mathrm{b}}$Initial log standard deviation for state-dependent exploration.} \\
\end{tabular}
\end{table}

\subsubsection{SkipPPO Trainer}\label{sec:skippPOTrainer}

The vastly different time horizons of the RWs and MTs, spanning two orders of magnitude, introduced a conflict with regard to the optimization targets and training hyperparameters. This introduced a pronounced credit assignment problem. The RWs have to focus on immediate rewards, as wrong RW commands can quickly spin up the satellite, whereas MTs have to consider long time horizons due to the small torques they generate. Further, the MTs are underactuated. Additionally, due to the torque difference, even single RW actions can cancel out hundreds of MT commands. Subsequently, using the naive training approach, it cannot always be determined which MT actions effectively contributed to a certain outcome and reward. To mitigate the credit assignment problem, PPO was modified to SkipPPO, allowing optional action freezing and time step skipping. When skip training is used, MT actions are frozen for $\mathrm{num\_skip}$ steps while RW actions continue to be sampled each step. Rewards from skipped steps are accumulated and assigned to the initial MT action, mitigating the credit assignment problem, effectively shortening the MT episode length to $\mathrm{episode\_length} / (\mathrm{num\_skip} + 1)$, and amplifying the effect of each individual MT action.

\subsubsection{Initial Base-Agent Training}\label{sec:training}

Training consisted of initial base-agent training and base-agent post-training. For the initial training only orbit parameters and inertia variations were considered. The RW subnetwork was trained first with its reward and hyperparameters, attaining random target attitudes. Then, the RW network was frozen while the MT network was trained to reach RW speeds of $\pm$500~\text{rpm}. This sequential training resolved conflicts between actuator requirements and reward functions (short term focus vs long term focus).

\subsubsection{Base-Agent Post-Training}\label{sec:finetuning}

To enable both subnetworks to adjust to each other, post-training was performed. The trainer received both RW and MT configurations simultaneously and switched between them during post-training. This allowed independent switching between RW- and MT subnetwork training within a single run, which improved coordination without retraining from scratch. For training stability, i.e., to ensure consistent convergence of the reinforcement learning process, the joint reward function from~{\eqref{eq:base_agent_combined_reward}} was used. The optimization thus always accounted for both actuators. Since the basic attitude control behavior had already been learned during the initial base-agent training, adaptation through post-training was simpler. With neither subnetwork deactivated and the reward function unchanged, the value network could reliably predict expected future rewards without being reinitialized. For post-training, therefore, the value network was carried over from the MT-training phase of the base-agent.

\subsection{Flight-Agent Design}\label{sec:flight_agent}

As briefly mentioned in Sections~\ref{sec:methodology} and~\ref{sec:base_agent}, while initially designed for momentum management, the first in-orbit test was carried out using only the RW subnetwork, since the logic for operating the attitude controller outside of the pass windows was not yet implemented, thus making the momentum management part non viable due to the maneuver length being significantly longer than the pass duration ($>$2000~s). As the in-orbit tests used active magnetic residual dipole moment compensation using the MTs, we have decided to use the base-agent, trained as previously described, as starting point for finetuning, instead of training a new controller from scratch. The reasoning behind this was, that as the RW subnetwork of the base-agent had to learn to accommodate for disturbances introduced by the MTs, it would transfer to the disturbances from the active residual dipole moment compensation. The following sections describe the applied changes to the base-agent.

\subsubsection{Safety Constraints}\label{safety_constraints}
Shortly after deployment and prior to the first test, the safety limits were tightened (see Section~\ref{sec:safetycage}). In particular the agent's maximum rotation rates were limited to 20~$^\circ$\text{/s} per axis. Further, the effective applied torque was constrained to 50~\text{rpm}/s per axis initially and shortly after relaxed to 100~\text{rpm}/s, corresponding to the imposed limit on the commanded torque. The maximum RW speeds were likewise initially set to [1000, 1000, 500]~\text{rpm} and increased to [1500, 1500, 700]~\text{rpm}. These limitations were implemented via clipping. Nevertheless, we tested the base-agent, resulting in the first successful demonstration of an AI-based attitude controller in orbit on October 30, 2025 during the pass from approximately 10:40 to 10:49 UTC. The initial attitude quaternion was $(-0.268, -0.851, -0.354, -0.280)$ with initial satellite body rates $(0.0227, -2.32, -1.74)$~$^\circ$\text{/s}, and the goal attitude quaternion was $(1, 0, 0, 0)$. Fig.~\ref{fig:maneuver_30.10.2025_10.40-10.50} shows the Yaw, Pitch and Roll Euler angles (corresponding to the X,Y, and Z satellite body axes, respectively), the commanded RW torques, the RW speeds and the satellite body rates for the maneuver, together with the times of the attitude control command (Maneuver Start) and when the goal attitude was attained with an error of $<$1$^\circ$ for all axes (Steady-State Start). Note that for all following figures the attitude components are given in Euler angles, which map vectors from the inertial reference frame into the satellite-fixed body frame; the satellite body rates are expressed in the satellite-fixed body frame, and the RW commands (torques) and speeds are expressed in their respective actuator frames; i.e., positive RW commands (in rpm/s) result in positive RW torque in the actuator frame, which result in positive body torque in the satellite body frame. All timestamps are given in UTC. Fig.~\ref{fig:maneuver_30.10.2025_10.40-10.50_steady_state} shows a detailed view of the steady-state period of the maneuver from Fig.~\ref{fig:maneuver_30.10.2025_10.40-10.50} and Table~\ref{tab:maneuver_30.10.2025_10.40-10.50_results} shows the numeric values for the minimum, maximum, mean and standard deviation of the yaw, pitch and roll angles for the steady-state period.

\Figure[t!](topskip=0pt, botskip=0pt, midskip=0pt)[width=0.99\linewidth]{djebk3.png}
{\textbf{First in-orbit maneuver of the base-agent on 2025-10-30: (a) attitude Euler angles, (b) commanded RW torques, (c) RW speeds, and (d) satellite body rates. The maneuver duration from the time step before the first observed command reaction (Maneuver Start; 10:43:24) until the error remained below $1^\circ$ for at least 15~s per axis (Steady-State Start; 10:45:18) resulted in a settling time of 114~s. Maximum wheel speeds were limited to [1000, 1000, 500]~\text{rpm} and applied torque to 50~\text{rpm}/s per axis (as opposed to the 100~\text{rpm}/s limit for the commanded torque) via clipping.}\label{fig:maneuver_30.10.2025_10.40-10.50}}

\Figure[t!](topskip=0pt, botskip=0pt, midskip=0pt)[width=0.99\linewidth]{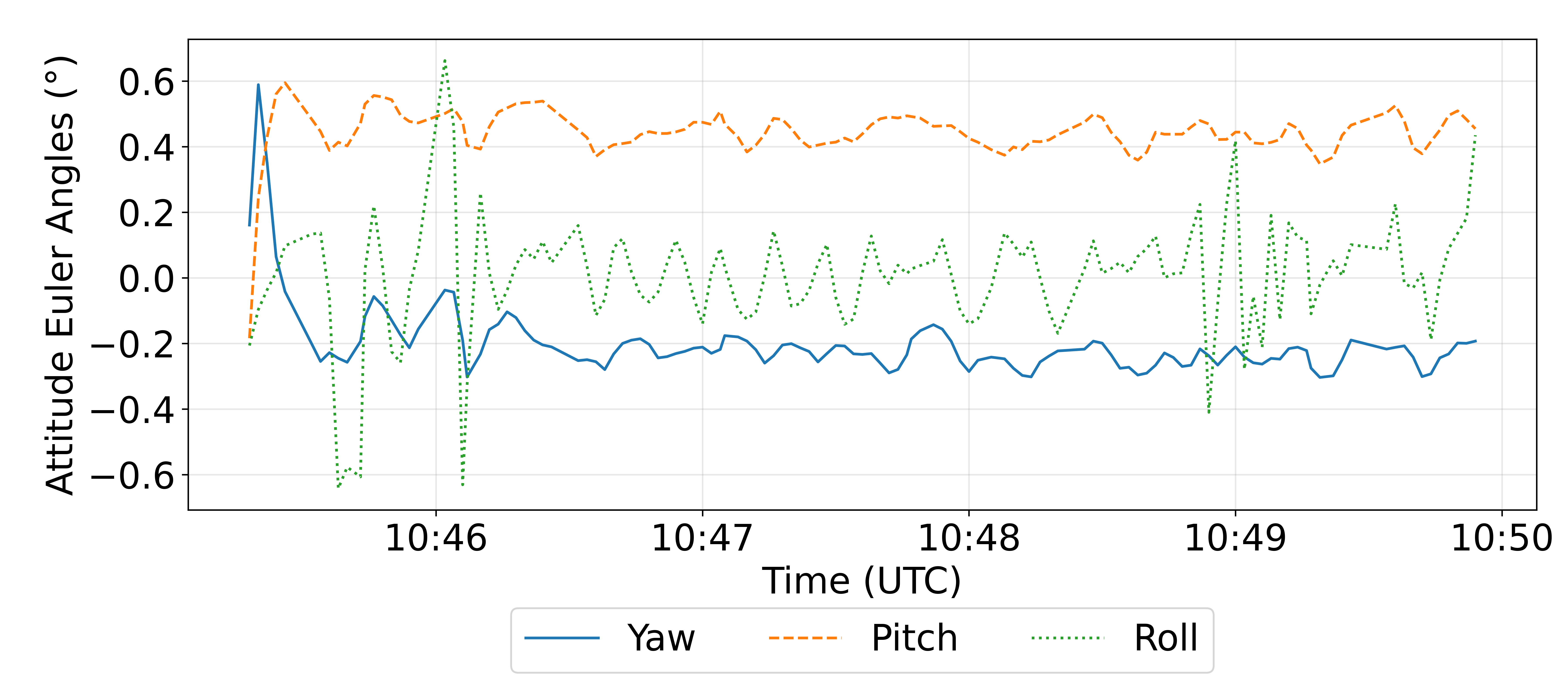}
{\textbf{Steady-state section of the first in-orbit maneuver of the base-agent from Fig.~\ref{fig:maneuver_30.10.2025_10.40-10.50}: Steady-state duration from 10:45:18 to 10:49:54 (276~s). The maximum wheel speeds were limited to [1000, 1000, 500]~\text{rpm} and the maximum torque to [50, 50, 50]~\text{rpm/s} via clipping.}\label{fig:maneuver_30.10.2025_10.40-10.50_steady_state}}

The maneuver start time is the first time step before RW commands are observable, accounting for potential misalignment between command execution and telemetry sampling, and preventing artificial reduction of maneuver duration. The start of the steady-state period is defined as the earliest time at which the attitude deviations across all three axes remain below $1^\circ$ for a continuous duration of at least 15~s. This threshold of $1^\circ$ was selected to match the target criterion from the reward function used during agent training. Consequently, achieving this level of precision in flight serves as a primary metric for successful Sim2Real transfer, demonstrating that the learned control policy generalizes effectively to the in-orbit environment. The reported in-orbit metrics characterize the performance of the controller under real flight conditions using the available on-board sensor data. All reported values are based on raw telemetry data. The telemetry data were captured at a sampling frequency of 0.5 Hz, exported from the InnoCube mission's Grafana dashboard, and are available in the GitHub repository at \url{https://github.com/kdjebko/lelar-in-orbit-data}. The RW subnetwork of the base-agent could operate under the clipping, although it was not part of the training, and was only (indirectly) affected by the 20~$^\circ$\text{/s} per axis limit. Exceeding this rate results in an instant controller shutdown by the Safety Cage. Since the base-agent was trained without a rotation rate constraint, the agent may exceed this limit when trying to attain distant attitudes. We decided to perform post-training for the base-agent considering these constraints for a fair performance evaluation. Table~\ref{tab:base_agent_safety_params} lists the relevant introduced safety-constraints and the appropriate settings used during the base-agent training.

\begin{table}[t!]
\caption{\textbf{Steady-State Error Statistics for the Maneuver of the LeLaR Base-Agent From Fig.~\ref{fig:maneuver_30.10.2025_10.40-10.50} From 10:45:18 to 10:49:54 (276~s).}}
\label{tab:maneuver_30.10.2025_10.40-10.50_results}
\setlength{\tabcolsep}{2.5pt}
\begin{tabular}{p{55pt} p{43pt} p{43pt} p{43pt} p{25pt}}
\hline
Axis & Min ($^\circ$) & Max ($^\circ$)& Mean ($^\circ$) & Std ($^\circ$) \\
\hline
Yaw   & -0.30 & 0.59 & -0.20 & 0.11 \\
Pitch & -0.18 & 0.59 & 0.44  & 0.08 \\
Roll  & \textbf{-0.64} & \textbf{0.66} & 0.00  & 0.18 \\
\hline
\end{tabular}
\end{table}

\begin{table}[t!]
\caption{\textbf{System Constraints and Safety Cage Limits (Single Values Apply to All Axes).}}
\label{tab:base_agent_safety_params}
\setlength{\tabcolsep}{3pt}
\begin{tabular}{p{60pt} p{75pt} p{80pt}}
\hline
Parameter (Max) & Base-Agent Training & Safety Limits\\
\hline
RW Speed & $1.64 \times 10^4$~\text{rpm} & $[1.5, 1.5, 0.7] \times 10^3$~\text{rpm}\\
[3pt]
RW Torque & $336.2$~\text{rpm/s} & $100$~\text{rpm/s}\\
& ($\approx 2 \times 10^{-3}$~\text{N}$\cdot$\text{m}) & \\
[3pt]
Sat. Body Rate & No limit imposed & $20~^\circ$\text{/s}\\
\hline
\multicolumn{3}{p{235pt}}{\scriptsize Note: RW values refer to $[X, Y, Z]$ body axes. All values refer to the physically possible maximum values (column Base-Agent Training) or the maximum allowed values (column Safety Limits).} \\
\end{tabular}
\end{table}

Additionally to the implementation of these changes for the flight-agent training, the RW min torque was marginally corrected to [1.5, 1.5, 1.5]~$\times$10$^{-5}$~\text{N}$\cdot$\text{m} after a cross-check with InnoCube. 
To reflect the standalone nature of the RW subnetwork the initial RW speeds were changed from $\pm$500~\text{rpm} to 0~\text{rpm} per axis for the training of the flight-agent. Further, the simulated boundary condition variations were increased for improved robustness and to make efficient use of the post-training run. The residual magnetic dipole moment compensation error was raised from $\pm$10\% to $\pm$25\% per axis and the initial satellite body rates were varied in the interval $[-5, 5]$~$^\circ$\text{/s} per axis. Table~\ref{tab:base_agent_additional_modifications} lists the modifications to the simulation scenario applied during flight-agent training.

\begin{table}[t!]
\caption{\textbf{Simulation Adjustments for the Flight-Agent Training (Single Values Apply to All Axes).}}
\label{tab:base_agent_additional_modifications}
\setlength{\tabcolsep}{3pt}
\begin{tabular}{p{75pt} p{70pt} p{70pt}}
\hline
Parameter & Base-Agent Training & Flight-Agent Training \\
\hline
Init. RW Speed & $\pm 500$~\text{rpm} & $0$~\text{rpm}\\
[3pt]
Init. Sat. Body Rate & $0$ $^\circ$\text{/s} & $U(-5, 5)$ $^\circ$\text{/s}\\
[3pt]
Dipole Comp. Error & $\pm 10\%$ & $\pm 25\%$ \\
\hline
\multicolumn{3}{p{235pt}}{\scriptsize Note: RW speed and body rates are defined per axis. The flight-agent training uses a more conservative modeled residual magnetic dipole compensation error to account for in-orbit uncertainties.} \\
\end{tabular}
\end{table}

\subsubsection{Flight-Agent Reward Function and Post-Training}\label{sec:effective_reward_function}
Based on the changed requirements regarding the satellite body rates, a modified reward function was employed. The Safety Cage limits from Table~\ref{tab:base_agent_additional_modifications} were incorporated into the simulation. While they were not negatively affecting the agent's performance in particular, they have been included here since it was not associated with further training cost. The modified RW reward function is shown in \eqref{eq:flight_agent_reward1}--\eqref{eq:flight_agent_reward9}.
\begin{equation}
    \mathrm{err\_att}_{t_i} = 1 - | \delta q_{0,t_i} |
    \label{eq:flight_agent_reward1}
\end{equation}
\begin{equation}
    \Delta \theta_{t_i} = \mathrm{err\_att}_{t_i} - \mathrm{err\_att}_{t_{i-1}}
    \label{eq:flight_agent_reward2}
\end{equation}
\begin{equation}
    \mathrm{p\_norm}_{t_i} = \| \mathbf{err\_rate}_{t_i} \|
    \label{eq:flight_agent_reward3}
\end{equation}
\begin{equation}
    \mathrm{p\_att}_{t_i} = 0.1 \cdot \mathrm{err\_att}_{t_i}
    \label{eq:flight_agent_reward4}
\end{equation}
\begin{equation}
    \mathrm{p\_ex}_{t_i} = \sum_{k=1}^{3} k_{\mathrm{ex}} \cdot \max(0, |\mathbf{err\_rate}_{t_i}[k]| - \mathrm{threshold})^2
\label{eq:flight_agent_reward5}
\end{equation}
\begin{equation}
    \mathrm{p\_smooth}_{t_i} = 
    k_{\mathrm{smooth}}\cdot
    \| \mathbf{a\_rw}_{t_i} - \mathbf{a\_rw}_{t_{i-1}} \|^2
    \label{eq:flight_agent_reward6}
\end{equation}
\begin{equation}
    \mathrm{p\_rates}_{t_i} = 0.5 \cdot \mathrm{p\_norm}_{t_i}^2
    \label{eq:flight_agent_reward7}
\end{equation}
\begin{equation}
r_{t_i} = 
\begin{cases}
1 + \frac{1}{\mathrm{p\_norm}_{t_i}+0.1} \\
\quad + \frac{1}{\mathrm{err\_att}_{t_i}\cdot 10^{5} + 0.1}, & \text{if } \mathrm{err\_att}_{t_i} < 3.8 {\cdot} 10^{-5} \\
e^{-\frac{\mathrm{err\_att}_{t_i}}{0.14}}, & \text{if } \mathrm{err\_att}_{t_i} < \mathrm{err\_att}_{t_{i-1}}\\
0.1 \cdot e^{-\frac{\Delta\theta_{t_i}}{0.14}} -1, & \text{if } \Delta \theta_{t_i} < \Delta \theta_{t_{i-1}} \\
e^{-\frac{\mathrm{err\_att}_{t_i}}{0.14}} - 2, & \text{else}
\end{cases}
\label{eq:flight_agent_reward8}
\end{equation}
\begin{equation}
\begin{split}
    \mathrm{reward\_rw}_{t_i} = r_{t_i} 
    & - \mathrm{p\_att}_{t_i} - \mathrm{p\_ex}_{t_i} \\
    & - \mathrm{p\_smooth}_{t_i} - \mathrm{p\_rates}_{t_i}
\end{split}
\label{eq:flight_agent_reward9}
\end{equation}

Here, $\mathbf{a\_rw}$ are the agent RW actions. Just like with the reward function used for training the base-agent, the expression $\delta q_{0,t_i}$ always equals $|\delta q_{0,t_i}|$ due to the mapping of the attitude error quaternion to its positive scalar representation as described in Section~\ref{sec:observationspace}. The main change compared to the RW reward function of the base-agent was the increase of the rotation rate regularization and the inclusion of the penalty terms $\mathrm{p\_ex}_{t_i}$ to punish rotation rates exceeding a certain threshold and $\mathrm{p\_smooth}_{t_i}$ to penalize bang-bang behavior to avoid reward hacking in conjunction with the rotation rate penalty. The parameter $\mathrm{threshold}$ was set to  10.0 ($^\circ$\text{/s}; as soft maximum for the desired satellite body rates, recommended by InnoCube operators, to provide a buffer to the Safety Cage limit of 20$^\circ$\text{/s} per axis), the parameter $k_{\mathrm{smooth}}$ to 50, and the parameter $k_{\mathrm{ex}}$ to 5.0. Further, the first term of $r_{t_i}$ was augmented to reward the agent stronger for lower steady state errors. For the fine tuning through post-training, the RW hyperparameters (post-training) from Table~\ref{tab:rw_mt_hyperparams} were used but with the number of rollout steps set to $2\times10^3$, the episode length set to 100 and the number of total time steps doubled to $1\times$10$^{9}$.

\subsection{Safety Cage and Integration}\label{sec:safetycageandintegration}
Two new modules were introduced to the InnoCube's ADCS software, that are necessary for the LeLaR execution on-board: a Safety Cage module, which ensures safe execution of control experiments, and the LeLaR controller module, which implements the developed AI agent.

\subsubsection{Safety Cage}\label{sec:safetycage}

In order to assure that unsafe states are not reached during attitude control maneuvers, a Safety Cage architecture as proposed in the ECSS Machine Learning Handbook~\cite{aiecsshandbook} is implemented. This construct uses a set of rules, to prevent the system from entering potentially dangerous states. This can be achieved through a set of safety interventions, which will be activated once a certain rule is triggered. Potentially critical failure modes for the InnoCube satellite have been identified using the proposed Failure Mode, Effects and Criticality analysis (FMECA). This analysis led to the main considered critical failure modes, described in the following.

The available angular momentum from the RWs could potentially result in very high satellite body rates (exceeding 20~$^\circ$\text{/s}) in case of a malfunctioning controller. High satellite body rates can lead to critical states by stressing the flexible antenna rods, degrading communication performance, and reducing the effectiveness of solar power tracking. Through the buildup of disturbance angular momentum in the RWs, high rotational body rates might also be induced when powering-off RWs, without prior desaturation. Incorrect commanded MT outputs might also lead to high rotational body rates, especially if applied for a prolonged period of time.

The Safety Cage is integrated into the existing ADCS software and complements the FDIR schemes with two components: a set of configurable safety rules defining hard limits on actuator and procedural parameters, and a monitoring module that enforces them. The main limits are RW torque and maximum RW speed. By restricting these, the achievable satellite body rates and maximum stored angular momentum, even under full RW saturation, are bounded. Operators can adjust limits via telecommand. During LeLaR experiments, maximum RW speed was the primary Safety Cage parameter. The total experiment duration was restricted to 15 minutes for safety reasons and power constraints.

The monitoring module runs independently and in parallel with the control loop. It receives all ADCS telemetry (e.g., raw sensor data, sensor and actuator states, and possible anomalies) and compares them against telecommanded rules. This ensures that software bugs, operator errors, or other unforeseen problems do not cause the system to transition into critical states. Monitored parameters include those defined as hard limits (e.g., RW speeds) as well as other telemetry, such as satellite rotational rates.

Exceeding a predefined limit causes the ADCS to disable the active controller and enter an idle state, where no actuator commands are issued. After a set period, the RWs are turned off to conserve power. Since InnoCube's energy budget is power-positive while tumbling (the solar-panels are mounted on each long side of the satellite), being in an idle state with deactivated actuators is both safe and efficient. After a safety trigger, it is the operator's responsibility to reset and reconfigure the system for new experiments. The main Safety Cage constraints used during the in-orbit experiments are described in Section~\ref{safety_constraints}.

\subsubsection{InnoCube LeLaR Integration}\label{sec:integration}

Due to the the modular ADCS software design, integrating the LeLaR controller module into the existing software was straightforward. The controller module treats the implementation as an input-output black box, accepting input data vectors (observations) and producing actuator torque requests (actions) as outputs. Internally, the LeLaR controller module is divided into two main components: the network inference engine and the network loader. The inference engine runs the forward pass of the neural network using the observation vector to generate actuator commands. The inference is based on a bare-metal implementation of the AI agent's policy networks. Without special emphasis on code optimization, the inference engine can execute a full forward pass in approximately 38~ms, which is sufficiently fast, considering the 1~Hz control loop frequency. The network loader allows changing the loaded network by reading network parameters from a file stored in the ADCS flash-storage. Each file contains the network architecture and the corresponding weights and biases obtained during training on ground. For the flight-agent, the fully trained and uncompressed network file resulted in a binary file of approximately 105~kB in size. Multiple network files can be stored on-board and loaded as needed using telecommands, allowing different trained networks to be evaluated efficiently during in-orbit experiments.

\section{Results}\label{sec:results}

In the following, first the discrepancies between the simulated and the real-world behavior of the physical satellite are discussed and the results from different in-orbit maneuvers of the AI-based attitude controller are presented. Two types of inertial pointing maneuvers were considered. The first type corresponds to the initial maneuver after activating the AI-based attitude controller and features a random initial attitude, i.e., the attitude at the time the control experiment starts. For subsequent maneuvers, corresponding to the second type, the current attitude was set to a specified value. For all maneuvers, the goal attitude was commanded to be $(1, 0, 0, 0)$. This choice is equivalent to keeping the current attitude as is and commanding specific goal attitudes, but it enabled quick visual comparability during ground passes based on the attitude quaternions from the raw housekeeping telemetry data available via the InnoCube mission's Grafana dashboard.

\subsection{Sim2Real discrepancies}\label{sec:sim2real_discrepancies}
The actual satellite behavior diverged from the simulation setup of the flight-agent. The reaction wheels showed a few peculiarities, which were not accounted for during training and which we demonstrate in Section~\ref{sec:rw_behaviour}. The RWs exhibit reduced reliability in the range of [-350, 350]~\text{rpm}. The measured wheel speeds can show sudden spikes of more than 150~\text{rpm}, creating a discrepancy between the measured and executed wheel speeds. Another issue is, that when activating the RWs at the experiment start, each wheel has a random dead time of up to approximately 20 seconds. It can happen, that e.g., the X-RW starts spinning immediately, whereas the Y- and Z-RW only start responding 20 seconds later. While this behavior was also not modeled in simulation, this too posed no critical issues for the AI agent. Additionally, it was expected to employ both sets of MTs for residual dipole moment compensation. This compensation was assumed and modeled in simulation. However the planned software update to enable simultaneous operation of both ADCS modules was scrapped due to safety reasons and hence only one set of MTs remained for active dipole moment compensation. While it would have been possible to collect in-orbit data and then adjust the simulation model of the RWs, we decided against it to show the ability of the AI agent to adapt to deviations from the expected boundary conditions in a zero-shot fashion. Regarding the residual dipole moment compensation, also no further adjustments to the simulation were made, as the simulated compensation error window of 25\% was large enough to include the single MT case. In general, post-launch simulation tuning is often infeasible and contradicts the objective of achieving robust generalization. The AI-based controller is expected to compensate these discrepancies out-of-the-box. Hence the only reason to refine the simulation at this point would be to investigate the maximal achievable performance, which is not the intention of this work.

\subsection{In-Orbit Results}\label{sec:inorbitresults}
In the following, we present the in-orbit experiments conducted with the InnoCube nanosatellite. We first demonstrate the effect of the RWs becoming trapped within a narrow rotation speed interval and illustrate a representative inertial pointing maneuver in the presence of RW dead time. Subsequently, we discuss two sets of experiments presenting the LeLaR flight-agent's in-orbit performance. We provide a reference using the corresponding simulated maneuvers and the in-orbit performance of the classical PD controller currently implemented on-board InnoCube. The first set of experiments consists of repeated executions of an identical maneuver, whereas the second experiment evaluates the performance for a predefined sequence of attitudes. The maneuver telemetry data (including the data of the first base-agent maneuver from Fig.~\ref{fig:maneuver_30.10.2025_10.40-10.50}) were captured at a sampling frequency of 0.5 Hz and are available in the GitHub repository at \url{https://github.com/kdjebko/lelar-in-orbit-data} as obtained via export from the InnoCube mission's Grafana dashboard.

The experimental results are assessed using steady-state performance metrics. This choice is motivated by the limited observability and repeatability of transient dynamics in the in-orbit environment, which are strongly influenced by unmodeled disturbances, actuator nonlinearities, and timing uncertainties, especially given the limited number of experiments that can be performed in-orbit. In contrast, steady-state behavior provides a robust and comparable measure of control performance, directly reflecting pointing accuracy and enabling a meaningful performance quantification in the presence of the aforementioned deviations and disturbances.

\subsubsection{In-Orbit RW Behavior}\label{sec:rw_behaviour}

The RWs exhibited certain behavioral characteristics impeding attitude control when being operated in low rpm ranges. This includes sudden spikes of the measured RW speeds, unresponsiveness to RW commands near 0~\text{rpm} and a random dead time after turning the RWs on.
Fig.~\ref{fig:mrw_jump} shows in-orbit data of an RW test from December 15, 2025. A sudden jump in the measured RW speeds by approximately 185 rpm is present at 21:48:54. As can be seen by the unaffected satellite attitude, this outlier is purely a measurement artifact.

\Figure[t!](topskip=0pt, botskip=0pt, midskip=0pt)[width=0.99\linewidth]{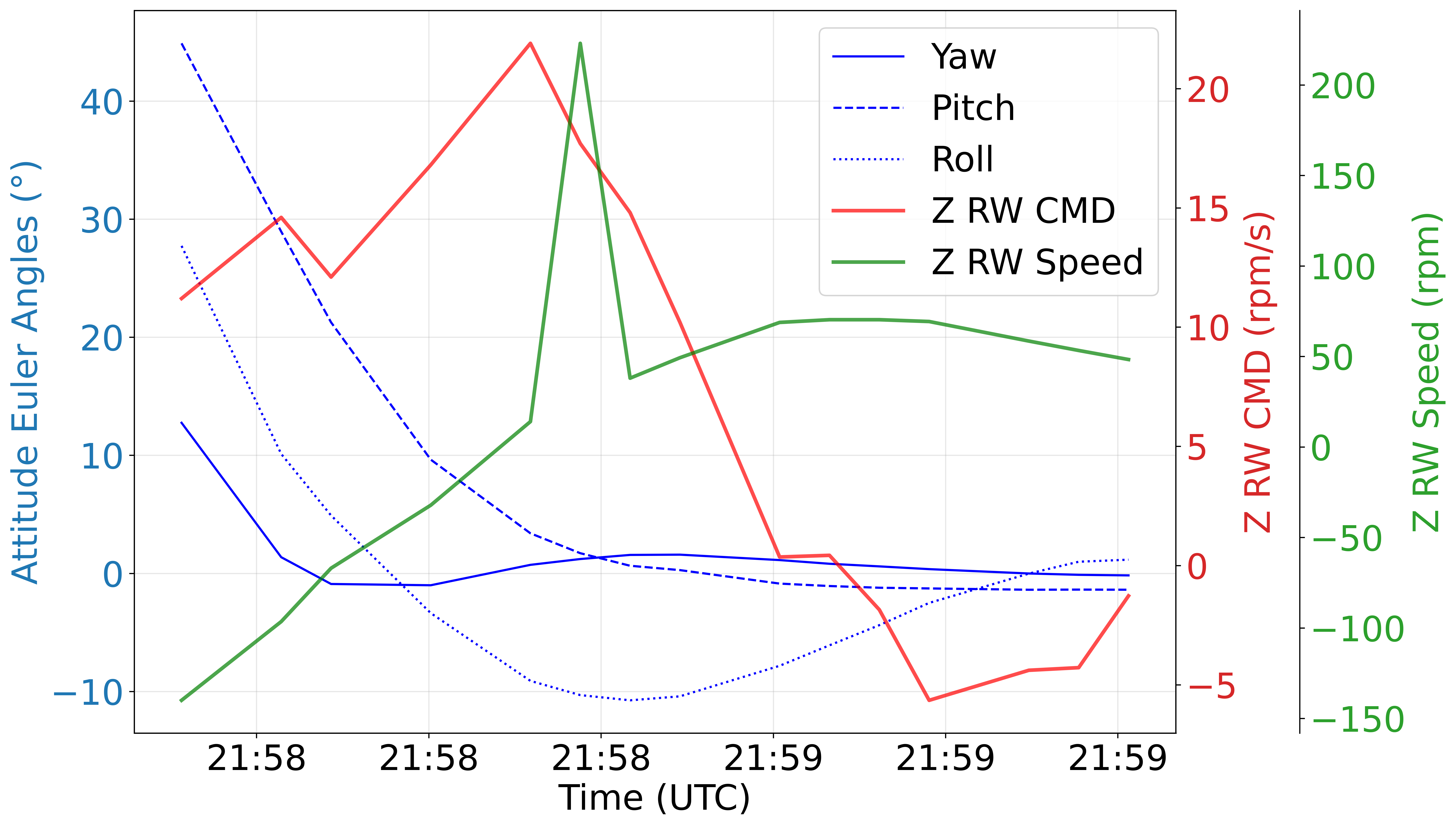}
{\textbf{In-orbit data showing a sudden spike in the measured Z-axis RW speed.}\label{fig:mrw_jump}}

Fig.~\ref{fig:maneuver_08.12.2025_22.19-22.25} shows rate estimator inaccuracies and unresponsiveness of the RWs from December 08, 2025. Around 20:23:20 and 22:23:38 the AI agent commands positive torques for the X-axis RW, which does not react for several seconds, and then suddenly spikes. This happens twice in a row and is responsible for delayed attainment of the steady-state.

\Figure[t!](topskip=0pt, botskip=0pt, midskip=0pt)[width=0.99\linewidth]{djebk6.png}
{\textbf{Attitude control maneuver of the LeLaR flight-agent on 2025-12-08: (a) attitude Euler angles, (b) commanded RW torques, (c) RW speeds, and (d) satellite body rates. The maneuver started at 22:21:50, and steady-state pointing was achieved at 22:24:02 (settling time: 132~s).}\label{fig:maneuver_08.12.2025_22.19-22.25}}

Fig.~\ref{fig:maneuver_08.12.2025_22.19-22.25_zoom} shows a detailed snapshot of this timeframe. It can be seen, that the X-axis RW does not respond to commanded torques, making it impossible to accurately maintain the goal attitude during this time. On the other hand, the AI agent maintains overall control and attains the steady-state reliably as soon as the RWs resume nominal operation.

\Figure[t!](topskip=0pt, botskip=0pt, midskip=0pt)[width=0.99\linewidth]{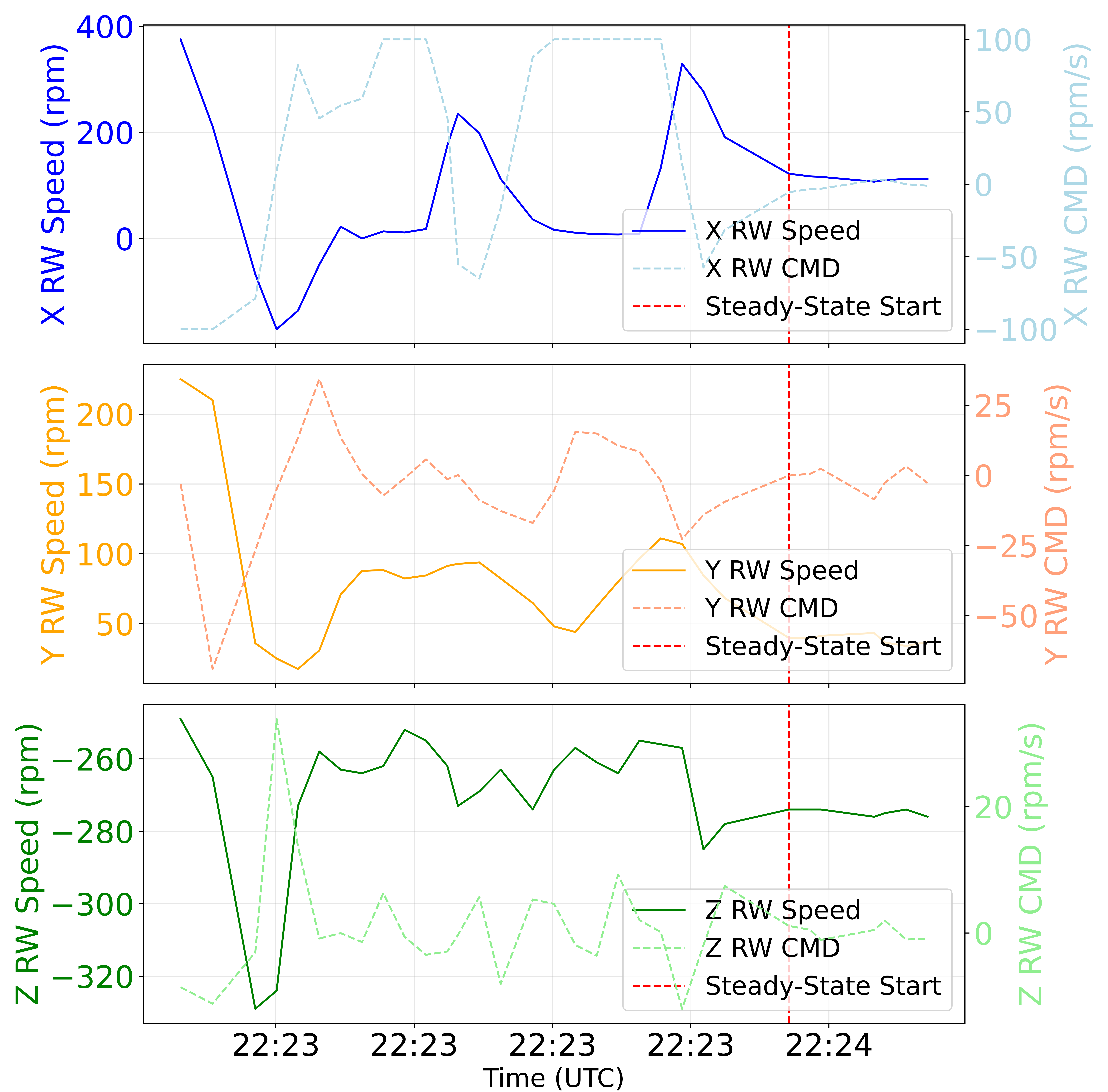}
{\textbf{Snapshot from the maneuver of Fig. \ref{fig:maneuver_08.12.2025_22.19-22.25}, illustrating the unresponsiveness of the X-RW towards commanded torques.} \label{fig:maneuver_08.12.2025_22.19-22.25_zoom}}

Another issue encountered was that the RWs exhibited a random dead time up to approximately 20~s after being turned on. During this time the RWs are unresponsive to commanded torques. Fig.~\ref{fig:maneuver_13.12.2025_11:29:21-11:30:19} shows an example maneuver from December 13, 2025 including RW dead time. For this maneuver, the goal attitude $(1, 0, 0, 0)$ was commanded. The initial attitude was $(-0.530, -0.0994, -0.418, 0.732)$ with an initial rotation rate of $(-0.140, -0.137, 6.16)$~$^\circ$\text{/s}.

\Figure[t!](topskip=0pt, botskip=0pt, midskip=0pt)[width=0.99\linewidth]{djebk8.png}
{\textbf{Attitude maneuver performance on 2025-12-13: (a) attitude Euler angles, (b) commanded RW torques, (c) RW speeds, and (d) satellite body rates. The maneuver started at 11:29:21 and the steady-state start was achieved at 11:30:19 (settling time: 58~s).}\label{fig:maneuver_13.12.2025_11:29:21-11:30:19}}

It can be seen from Fig.~\ref{fig:maneuver_13.12.2025_11:29:21-11:30:19} (c), that the RWs do not react immediately, instead the Z-axis RW reacts first, followed by the Y-axis RW and finally the X-axis RW after a random delay that cannot be anticipated in advance. Table~\ref{tab:maneuver_13.12.2025_11:29:21-11:30:19_steady_state_zoomed} quantifies the steady-state period. As with Table~\ref{tab:maneuver_30.10.2025_10.40-10.50_results}, the start of the steady-state period is defined as the time where the attitude deviations of all three axes settle below $1^\circ$ each for at least 15~s. Fig.~\ref{fig:maneuver_13.12.2025_11.28-11.34_steady_state} visualizes the steady-state-period of the maneuver from Fig.~\ref{fig:maneuver_13.12.2025_11:29:21-11:30:19}.

\begin{table}[t!]
\centering
\caption{\textbf{Steady-State Error Statistics for the Maneuver of Fig.~\ref{fig:maneuver_13.12.2025_11:29:21-11:30:19} From 11:30:19 to 11:31:28 (69~s).}}\label{tab:maneuver_13.12.2025_11:29:21-11:30:19_steady_state_zoomed}
\setlength{\tabcolsep}{3pt}
\begin{tabular}{p{55pt} p{43pt} p{43pt} p{43pt} p{25pt}}
\hline
Axis & Min ($^\circ$) & Max ($^\circ$)& Mean ($^\circ$) & Std ($^\circ$) \\
\hline
Yaw     &  -0.12 &   \textbf{0.38} &   0.03 &   0.09 \\
Pitch   &  -0.08 &   0.33 &   0.04 &   0.07 \\
Roll    &  \textbf{-0.30} &   0.04 &  -0.09 &   0.07 \\
\hline
\end{tabular}
\end{table}

\Figure[t!](topskip=0pt, botskip=0pt, midskip=0pt)[width=0.99\linewidth]{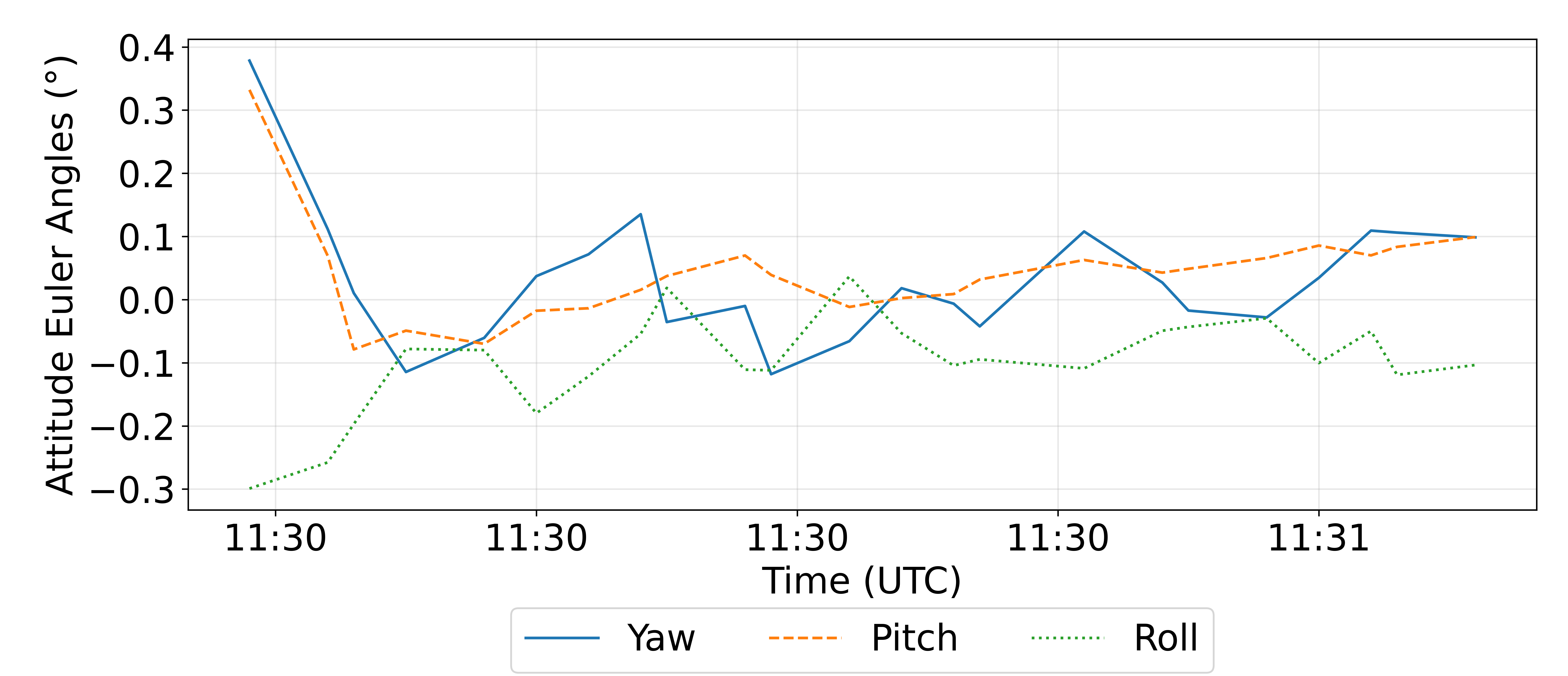}
{\textbf{Steady-state section of the maneuver of the flight-agent from Fig. \ref{fig:maneuver_13.12.2025_11:29:21-11:30:19}. Steady-state period: 11:30:19 to 11:31:28 (69~s).}\label{fig:maneuver_13.12.2025_11.28-11.34_steady_state}}

\subsubsection{Performance for Repeated Maneuvers}\label{sec:repeated_maneuvers}
Experiments using command lists were performed where a rotation from $(-0.5, -0.5, -0.5, -0.5)$ to $(1, 0, 0, 0)$ was conducted for the LeLaR flight-agent in-orbit and the PD controller in-orbit of InnoCube. Additionally, the LeLaR in-orbit maneuver was simulated to provide a Sim2Real reference. The default PD controller was employed in its strictly as-is state, as it was readily available as the pre-integrated default controller of InnoCube. Rather than a direct performance comparison, these results provide a contextual reference outlining the satellite's default operating behavior.

It should be noted that the resulting in-orbit metrics represent a conservative estimate of the controller's performance. Since the telemetry includes (remaining) uncompensated gyroscope bias and stochastic measurement noise, the control accuracy is subject to sensor uncertainty, which is difficult to determine precisely without an external reference. For the first maneuver, the current attitude and satellite body rates remained unmodified and the goal attitude was set to $(1, 0, 0, 0)$. For all subsequent maneuvers, the procedure was to set the current attitude to $(-0.5, -0.5, -0.5, -0.5)$ and then to command $(1, 0, 0, 0)$ as goal attitude. For these maneuvers, the initial satellite body rates were close to 0. A delay of 120~s between maneuvers was used, allowing for seven maneuvers to be completed in the maximum experiment timeframe of 15 minutes, which was set due to safety and power constraints for unsupervised experiments. Fig.~\ref{fig:maneuver_17.12.2025_20.46-21.01} shows the attitude quaternions, commanded RW torques, RW speeds and satellite body rates, along with the maneuver start and settling-time start times for a LeLaR flight-agent experiment from December 17, 2025. The initial attitude for the first maneuver was $(0.973, -0.0967, 0.105, -0.180)$ and the initial satellite body rates were $(-0.974, 0.917, -1.46)$~$^\circ$\text{/s}. Table~\ref{tab:lelar_stats_17.12.2025_20.46-21.01} quantifies the steady-state behavior of these maneuvers.

\Figure[t!](topskip=0pt, botskip=0pt, midskip=0pt)[width=0.99\linewidth]{djebk10.png}
{\textbf{In-orbit flight telemetry for seven sequential maneuvers (M1--M7) performed by the LeLaR flight-agent on 2025-12-17: (a) attitude Euler angles, (b) commanded RW torques, (c) RW speeds, and (d) satellite body rates. Maneuver intervals (Maneuver Start--Steady-State Start; settling-time): M1 (20:46:21--20:47:05; 44~s), M2 (20:48:19--20:49:51; 92~s), M3 (20:50:19--20:51:25; 66~s), M4 (20:52:19--20:53:03; 44~s), M5 (20:54:19--20:55:55; 96~s), M6 (20:56:19--20:57:33; 74~s), and M7 (20:58:19--20:59:03; 44.0~s).}\label{fig:maneuver_17.12.2025_20.46-21.01}}

\begin{table}[t!]
\caption{\textbf{Steady-State Error Statistics for the Maneuvers M1--M7 of the LeLaR Flight-Agent From Fig.~\ref{fig:maneuver_17.12.2025_20.46-21.01} on 2025-12-17.}} \label{tab:lelar_stats_17.12.2025_20.46-21.01}
\setlength{\tabcolsep}{3pt}
\begin{tabular}{p{43pt}p{34pt}p{34pt}p{34pt}p{34pt}p{26pt}}
\hline
ID (duration)& Comp. & Min ($^\circ$) & Max ($^\circ$)& Mean ($^\circ$) & Std ($^\circ$) \\
\hline
\multirow{3}{*}{M1 (74 s)} & Yaw   & -0.06 &  0.11 &  0.03 &  0.04 \\
                           & Pitch & -0.06 &  0.18 &  0.11 &  0.05 \\
                           & Roll  & \textbf{-0.43} &  \textbf{0.43} &  0.03 &  0.20 \\ \hline
\multirow{3}{*}{M2 (28 s)} & Yaw   & \textbf{-0.52} &  \textbf{0.74} &  0.04 &  0.30 \\
                           & Pitch & -0.47 &  0.17 & -0.00 &  0.18 \\
                           & Roll  & -0.35 &  0.14 & -0.02 &  0.15 \\ \hline
\multirow{3}{*}{M3 (54 s)} & Yaw   & -0.04 &  0.29 &  0.08 &  0.09 \\
                           & Pitch & -0.13 &  0.48 &  0.12 &  0.19 \\
                           & Roll  & \textbf{-0.51} &  \textbf{0.63} & -0.04 &  0.30 \\ \hline
\multirow{3}{*}{M4 (76 s)} & Yaw   & -0.39 &  0.11 &  0.02 &  0.09 \\
                           & Pitch & -0.28 &  \textbf{0.89} &  0.13 &  0.17 \\
                           & Roll  & \textbf{-0.91} &  0.63 & -0.04 &  0.32 \\ \hline
\multirow{3}{*}{M5 (24 s)} & Yaw   & \textbf{-0.47} &  0.15 & -0.05 &  0.22 \\
                           & Pitch & -0.05 &  \textbf{0.30} &  0.13 &  0.11 \\
                           & Roll  & -0.18 &  0.22 & -0.01 &  0.12 \\ \hline
\multirow{3}{*}{M6 (46 s)} & Yaw   & -0.05 &  0.41 &  0.05 &  0.12 \\
                           & Pitch & \textbf{-0.31} &  \textbf{0.50} &  0.07 &  0.25 \\
                           & Roll  & -0.18 &  0.18 & -0.06 &  0.09 \\ \hline
\multirow{3}{*}{M7 (74 s)} & Yaw   & -0.07 &  0.17 &  0.04 &  0.05 \\
                           & Pitch & -0.04 &  0.14 &  0.03 &  0.04 \\
                           & Roll  & \textbf{-0.42} &  \textbf{0.27} & -0.12 &  0.16 \\ \hline
\end{tabular}
\end{table}

Using the initial attitude and initial rates, the maneuvers from Fig.~\ref{fig:maneuver_17.12.2025_20.46-21.01} were simulated to provide a reference for the differences between simulated and actual in-orbit boundary conditions that influence the Sim2Real transfer. The simulated data were downsampled in order to match the sampling frequency of the in-orbit data of 0.5~Hz. As mentioned in Section~\ref{sec:innocube_adcs}, only relative attitudes are used in the in-orbit experiments. Therefore, initial conditions (mainly the magnetic field vector) might differ in simulation, which is negligible for this analysis due to the reaction wheels providing orders of magnitude higher control torque than the expected magnetic disturbances. The maneuver trajectories can be seen in Fig.~\ref{fig:sim_maneuver_0_5}. The corresponding steady-state metrics are shown in Table~\ref{tab:sim_stats_0_5}.

\Figure[t!](topskip=0pt, botskip=0pt, midskip=0pt)[width=0.99\linewidth]{djebk11.png}
{\textbf{Simulated maneuvers modeled after the in-orbit maneuvers from Fig.~\ref{fig:maneuver_17.12.2025_20.46-21.01}: (a) attitude Euler angles, (b) commanded RW torques, (c) RW speeds, and (d) satellite body rates. Maneuver intervals (Maneuver Start--Steady-State Start; settling-time): M1 (20:46:21--20:46:35; 14~s), M2 (20:48:19--20:48:57; 38~s), M3 (20:50:19--20:50:57; 38~s), M4 (20:52:19--20:52:55; 36~s), M5 (20:54:19--20:54:57; 38~s), M6 (20:56:19--20:56:57; 38~s), and M7 (20:58:19--20:58:55; 36~s).}\label{fig:sim_maneuver_0_5}}

\begin{table}[t!]
\caption{\textbf{Steady-State Error Statistics for the Simulated Maneuvers M1--M7 From Fig.~\ref{fig:sim_maneuver_0_5}, Modeling the In-Orbit Maneuvers of the LeLaR Flight-Agent From Fig.~\ref{fig:maneuver_17.12.2025_20.46-21.01} on 2025-12-17.}} \label{tab:sim_stats_0_5}
\setlength{\tabcolsep}{3pt}
\begin{tabular}{p{43pt}p{34pt}p{34pt}p{34pt}p{34pt}p{26pt}}
\hline
ID (duration)& Comp. & Min ($^\circ$) & Max ($^\circ$)& Mean ($^\circ$) & Std ($^\circ$) \\
\hline
\multirow{3}{*}{M1 (104 s)} & Yaw   & -0.65 &  0.27 & -0.02 &  0.11 \\
                            & Pitch & -0.44 &  0.31 &  0.01 &  0.11 \\
                            & Roll  & \textbf{-0.74} &  \textbf{0.87} & -0.02 &  0.19 \\
\hline
\multirow{3}{*}{M2 (82 s)}  & Yaw   & -0.10 &  \textbf{0.43} &  0.10 &  0.09 \\
                            & Pitch & \textbf{-0.51} &  0.15 &  0.00 &  0.14 \\
                            & Roll  & -0.04 &  0.42 &  0.06 &  0.08 \\
\hline
\multirow{3}{*}{M3 (82 s)}  & Yaw   & \textbf{-0.63} &  0.04 & -0.07 &  0.11 \\
                            & Pitch & -0.13 &  0.14 &  0.05 &  0.06 \\
                            & Roll  & -0.01 &  \textbf{0.30} &  0.07 &  0.06 \\
\hline
\multirow{3}{*}{M4 (84 s)}  & Yaw   & -0.48 &  0.16 &  0.01 &  0.09 \\
                            & Pitch & -0.48 &  0.15 & -0.03 &  0.11 \\
                            & Roll  & \textbf{-0.54} &  \textbf{0.69} & -0.01 &  0.16 \\
\hline
\multirow{3}{*}{M5 (82 s)}  & Yaw   & -0.21 &  \textbf{0.30} &  0.04 &  0.09 \\
                            & Pitch & \textbf{-0.54} &  0.08 & -0.04 &  0.14 \\
                            & Roll  & -0.29 &  0.17 & -0.17 &  0.08 \\ 
\hline
\multirow{3}{*}{M6 (82 s)}  & Yaw   & \textbf{-0.67} &  0.14 &  0.01 &  0.12 \\
                            & Pitch & -0.20 &  0.11 &  0.01 &  0.07 \\
                            & Roll  & -0.23 &  \textbf{0.19} & -0.04 &  0.09 \\ 
\hline
\multirow{3}{*}{M7 (84 s)}  & Yaw   & \textbf{-0.67} &  0.11 & -0.09 &  0.11 \\
                            & Pitch & -0.66 &  0.01 & -0.11 &  0.12 \\
                            & Roll  & -0.53 &  \textbf{0.79} &  0.03 &  0.19 \\ 
\hline
\end{tabular}
\end{table}

The same experiment was conducted using the default PD controller of InnoCube. The PD controller serves as the original InnoCube controller and was manually tuned in simulation, prior to, and independently of, the LeLaR AI agent training (see Section~\ref{sec:innocube_pd}). The same set of seven maneuvers was commanded as those performed by the LeLaR flight-agent. Each time the initial attitude was set to $(-0.5, -0.5, -0.5, -0.5)$ and the goal attitude to $(1, 0, 0, 0)$. Given a goal steady-state requirement of $1^\circ$ for each axis, the goal steady-state was only attained during the PD maneuvers 3, 5, and 6. For the first maneuver, the initial attitude was $(0.893, -0.0282, -0.0269, 0.448)$ and the initial satellite body rates were $(-0.348, -0.244, 4.09)$~$^\circ$\text{/s}. Fig.~\ref{fig:pd_maneuver_15.12.2025_21.50-22.05} visualizes the PD maneuvers from the experiment conducted on December 15, 2025. Table~\ref{tab:pd_stats_15.12.2025_21.50-22.05} shows the steady-state statistics for the successful PD maneuvers 3, 5, and 6.

\Figure[t!](topskip=0pt, botskip=0pt, midskip=0pt)[width=0.99\linewidth]{djebk12.png}
{\textbf{In-orbit flight telemetry for seven sequential maneuvers (M1--M7) using the default PD controller on 2025-12-15: (a) attitude Euler angles, (b) commanded RW torques, (c) RW speeds, and (d) satellite body rates. Maneuver intervals with achieved steady-state (Maneuver Start--Steady-State Start; settling-time): M3 (21:54:18--21:55:24; 66~s), M5 (21:58:18--21:59:30; 72~s), and M6 (22:00:18--22:01:32; 74~s). Remaining maneuver start times (Maneuver Start): M1 (21:50:18), M2 (21:52:18), M4 (21:56:18), and M7 (22:02:18).}\label{fig:pd_maneuver_15.12.2025_21.50-22.05}}

\begin{table}[t!]
\caption{\textbf{Steady-State Error Statistics for the Successful Maneuvers M3, M5, and M6 of the PD Controller From Fig.~\ref{fig:pd_maneuver_15.12.2025_21.50-22.05} on 2025-12-15.}}\label{tab:pd_stats_15.12.2025_21.50-22.05}
\setlength{\tabcolsep}{3pt}
\begin{tabular}{p{43pt}p{34pt}p{34pt}p{34pt}p{34pt}p{26pt}}
\hline
ID (duration)& Comp. & Min ($^\circ$) & Max ($^\circ$)& Mean ($^\circ$) & Std ($^\circ$) \\
\hline
\multirow{3}{*}{M3 (54 s)} & Yaw   & -0.01 &  0.34 &  0.22 &  0.10 \\
                           & Pitch & -0.33 &  0.23 & -0.00 &  0.20 \\
                           & Roll  & \textbf{-0.43} &  \textbf{2.58} & 0.75 &  0.92 \\ \hline
\multirow{3}{*}{M5 (48 s)} & Yaw   & -0.23 &  0.32 &  0.12 &  0.18 \\
                           & Pitch & \textbf{-0.75} & -0.22 & -0.50 &  0.17 \\
                           & Roll  &  0.18 &  \textbf{0.99} &  0.54 &  0.23 \\ \hline
\multirow{3}{*}{M6 (46 s)} & Yaw   & -0.20 &  0.28 & -0.01 &  0.15 \\
                           & Pitch & -0.29 &  0.20 & -0.14 &  0.18 \\
                           & Roll  & \textbf{-1.97} &  \textbf{0.79} & -0.88 &  0.74 \\ \hline
\end{tabular}
\end{table}

In order for the PD controller to attain steady-state during all maneuvers, the attitude error threshold needed to be adjusted. It was relaxed in $0.1^\circ$ increments until the steady-state was achieved for all maneuvers. This resulted in a final threshold of $1.8^\circ$ per axis. The corresponding steady-state metrics are shown in Table~\ref{tab:pd_stats_15.12.2025_21.50-22.05_relaxed}.

\begin{table}[t!]
\caption{\textbf{Steady-State Error Statistics for PD Maneuvers M1--M7 From Fig.~\ref{fig:pd_maneuver_15.12.2025_21.50-22.05}, Based on a Relaxed Steady-State Threshold of $1.8^\circ$ Per Axis to Allow the PD Controller to Reach Steady-State During All Maneuvers.}}\label{tab:pd_stats_15.12.2025_21.50-22.05_relaxed}
\setlength{\tabcolsep}{3pt}
\begin{tabular}{p{43pt}p{34pt}p{34pt}p{34pt}p{34pt}p{26pt}}
\hline
ID (duration)& Comp. & Min ($^\circ$) & Max ($^\circ$)& Mean ($^\circ$) & Std ($^\circ$) \\
\hline
\multirow{3}{*}{M1 (30 s)}  & Yaw   & -0.05 &  \textbf{0.20} &  0.08 &  0.09 \\
                            & Pitch & -0.27 & -0.04 & -0.15 &  0.07 \\
                            & Roll  & \textbf{-1.80} & -1.51 & -1.62 &  0.08 \\ \hline
\multirow{3}{*}{M2 (74 s)}  & Yaw   & -0.09 &  0.24 &  0.15 &  0.06 \\
                            & Pitch & -0.73 &  0.25 & -0.23 &  0.26 \\
                            & Roll  & \textbf{-1.14} &  \textbf{1.89} &  1.31 &  0.58 \\ \hline
\multirow{3}{*}{M3 (62 s)}  & Yaw   & -0.01 &  0.34 &  0.23 &  0.10 \\
                            & Pitch & -0.33 &  0.23 & -0.00 &  0.19 \\
                            & Roll  & \textbf{-1.68} &  \textbf{2.58} &  0.61 &  1.06 \\ \hline
\multirow{3}{*}{M4 (64 s)}  & Yaw   & -1.11 &  0.28 & -0.08 &  0.33 \\
                            & Pitch & -1.32 &  0.45 & -0.05 &  0.35 \\
                            & Roll  & \textbf{-1.51} &  \textbf{1.88} &  0.90 &  1.06 \\ \hline
\multirow{3}{*}{M5 (52 s)}  & Yaw   & -0.29 &  0.32 &  0.09 &  0.21 \\
                            & Pitch & \textbf{-0.88} & -0.22 & -0.52 &  0.18 \\
                            & Roll  &  0.18 &  \textbf{1.65} &  0.61 &  0.33 \\ \hline
\multirow{3}{*}{M6 (60 s)}  & Yaw   & -0.20 &  0.45 &  0.09 &  0.22 \\
                            & Pitch & -0.29 &  0.20 & -0.07 &  0.19 \\
                            & Roll  & \textbf{-1.97} &  \textbf{1.52} & -0.58 &  0.95 \\ \hline
\multirow{3}{*}{M7 (90 s)}  & Yaw   & -0.66 &  1.54 & -0.06 &  0.36 \\
                            & Pitch & -1.47 &  0.46 & -0.12 &  0.62 \\
                            & Roll  & \textbf{-1.80} &  \textbf{1.84} &  0.77 &  1.28 \\ \hline
\end{tabular}
\end{table}

Comparing the simulated and in-orbit maneuvers of the flight agent shows an increase in maneuver durations, indicating differences in the transient dynamics between the modeled and actual satellite. Despite these discrepancies, the agent achieves similar steady-state pointing accuracy in both simulation and in-orbit operations, reliably remaining within the $1^\circ$ band around the target attitude, given as training objective. This demonstrates that, although the transient response differs, the agent successfully generalizes to the real environment, achieving the Sim2Real transfer. Although the default PD controller exhibits smoother control trajectories, its steady-state error is larger, reflecting its original design target, namely meeting the satellite's primary mission requirements. As the PD controller was operated in its nominal state without post-launch fine-tuning, its performance serves as a characterization of the satellite's nominal hardware response. In this context, the AI-based controller, trained entirely in simulation, demonstrates a successful zero-shot transfer by achieving high-precision steady-state performance under actual in-orbit conditions. This result validates the capacity of the DRL AI agent to solve the attitude control problem through autonomous interaction with a simulator, achieving its goal even when faced with the in-orbit discrepancies and unmodeled dynamics discussed in Section~\ref{sec:sim2real_discrepancies}.

\subsubsection{Performance for Diverse Maneuvers}
Experiments were conducted using a list of quaternions for both the LeLaR flight-agent and the PD controller. Similar to the previous experiments, as goal attitudes the quaternion $(1, 0, 0, 0)$ was commanded. While the first maneuver utilized the satellite's current state, subsequent maneuvers began from specific quaternions (see Table~\ref{tab:maneuver_list}) with near-zero body rates. The delay between two consecutive maneuvers was increased to 150~s so that distant attitudes could be attained and steady-state periods recorded. Since the experiment duration was restricted to 15 minutes due to safety and power constraints, only six maneuvers could be performed with this delay. For the first maneuver, the initial attitude quaternion was $(0.906, -0.0976, 0.0253, -0.410)$ and the initial satellite body rates $(-0.822, -0.0569, -3.53)$~$^\circ$\text{/s}. Fig.~\ref{fig:maneuver_15.12.2025_09.31-09.49} shows the attitude quaternions, commanded RW torques, RW speeds and satellite body rates, together with the maneuver start and steady-state start times for the LeLaR experiment from December 15, 2025. Table~\ref{tab:lelar_stats_15.12.2025_09.31-09.49} quantifies the steady-state behavior of the maneuvers.
\begin{table}[t!]
\caption{\textbf{List of Initial Commanded Quaternions.}}
\label{tab:maneuver_list}
\setlength{\tabcolsep}{3pt}
\begin{tabular}{p{78pt}p{150pt}}
\hline
Maneuver & Commanded Quaternion $(q_0, q_1, q_2, q_3)$ \\
\hline
M1 & variable init. attitude/rates (see Section~\ref{sec:results})\\
M2 & (0.354, 0.612, -0.354, -0.612) \\
M3 & (0.008, -0.087, -0.087, 0.992) \\
M4 & (-0.5, -0.5, -0.5, -0.5) \\
M5 & (-0.16, 0.066, -0.377, -0.91) \\
M6 & (0.0, -0.009, -0.009, 1.0) \\
\hline
\end{tabular}
\end{table}

\Figure[t!](topskip=0pt, botskip=0pt, midskip=0pt)[width=0.99\linewidth]{djebk13.png}
{\textbf{In-orbit flight telemetry for six sequential maneuvers (M1--M6) performed by the LeLaR flight-agent on 2025-12-15, 2025: (a) attitude Euler angles, (b) commanded RW torques, (c) RW speeds, and (d) satellite body rates. Maneuver intervals (Maneuver Start--Steady-State Start; settling-time): M1 (09:31:12--09:32:50; 98~s), M2 (09:33:40--09:34:58; 78~s), M3 (09:36:12--09:37:54; 102~s), M4 (09:38:40--09:39:32; 52~s), M5 (09:41:12--09:43:16; 124~s), and M6 (09:43:40--09:44:46; 66~s).}\label{fig:maneuver_15.12.2025_09.31-09.49}}

\begin{table}[t!]
\caption{\textbf{Steady-State Error Statistics for the Maneuvers M1--M6 of the LeLaR Flight-Agent From Fig.~\ref{fig:maneuver_15.12.2025_09.31-09.49} on 2025-12-15.}}\label{tab:lelar_stats_15.12.2025_09.31-09.49}
\setlength{\tabcolsep}{3pt}
\begin{tabular}{p{43pt}p{34pt}p{34pt}p{34pt}p{34pt}p{26pt}}
\hline
ID (duration)& Comp. & Min ($^\circ$) & Max ($^\circ$)& Mean ($^\circ$) & Std ($^\circ$) \\
\hline
\multirow{3}{*}{M1 (50 s)}  & Yaw   & -0.34 &  0.21 & -0.05 &  0.15 \\
                            & Pitch & -0.24 &  \textbf{0.30} &  0.00 &  0.12 \\
                            & Roll  & \textbf{-0.39} &  0.27 &  0.01 &  0.20 \\ \hline
\multirow{3}{*}{M2 (74 s)}  & Yaw   & -0.03 &  \textbf{0.15} &  0.07 &  0.05 \\
                            & Pitch & -0.06 &  0.09 &  0.02 &  0.04 \\
                            & Roll  & \textbf{-0.38} &  0.07 & -0.09 &  0.09 \\ \hline
\multirow{3}{*}{M3 (46 s)}  & Yaw   & -0.05 &  0.34 &  0.02 &  0.09 \\
                            & Pitch & -0.16 &  0.09 &  0.00 &  0.07 \\
                            & Roll  & \textbf{-0.86} &  \textbf{0.57} &  0.03 &  0.29 \\ \hline
\multirow{3}{*}{M4 (100 s)} & Yaw   & -0.20 &  0.13 & -0.02 &  0.07 \\
                            & Pitch & \textbf{-0.78} &  \textbf{0.17} & -0.02 &  0.13 \\
                            & Roll  & -0.19 &  0.15 &  0.01 &  0.08 \\ \hline
\multirow{3}{*}{M5 (24 s)}  & Yaw   & -0.28 &  0.04 & -0.08 &  0.10 \\
                            & Pitch & -0.09 &  0.68 &  0.21 &  0.27 \\
                            & Roll  & \textbf{-0.65} &  \textbf{0.88} &  0.09 &  0.37 \\ \hline
\multirow{3}{*}{M6 (84 s)}  & Yaw   & -0.39 &  0.38 & -0.06 &  0.14 \\
                            & Pitch & -0.18 &  0.42 &  0.00 &  0.12 \\
                            & Roll  & \textbf{-0.67} &  \textbf{0.45} & -0.03 &  0.21 \\ \hline
\end{tabular}
\end{table}

As with the previous set of experiments, the in-orbit maneuvers were simulated with initial attitude and rates taken from the telemetry data. Fig.~\ref{fig:sim_maneuver_quaternion_list} shows the simulated trajectories, and Table~\ref{tab:sim_stats_quaternion_list} summarizes the steady-state metrics.

\Figure[t!](topskip=0pt, botskip=0pt, midskip=0pt)[width=0.99\linewidth]{djebk14.png}
{\textbf{Simulated maneuvers modeled after the in-orbit maneuvers from Fig.~\ref{fig:maneuver_15.12.2025_09.31-09.49}: (a) attitude Euler angles, (b) commanded RW torques, (c) RW speeds, and (d) satellite body rates. Maneuver intervals (Maneuver Start--Steady-State Start; settling-time): M1 (09:31:12--09:31:38; 26~s), M2 (09:33:40--09:35:16; 96~s), M3 (09:36:10--09:36:58; 48~s), M4 (09:38:40--09:39:18; 38~s), M5 (09:41:10--09:42:00; 50~s), and M6 (09:43:40--09:44:24; 44~s).}\label{fig:sim_maneuver_quaternion_list}}

\begin{table}[t!]
\caption{\textbf{Steady-State Error Statistics for the Simulated Maneuvers M1--M6 From Fig.~\ref{fig:sim_maneuver_quaternion_list}, Modeling the In-Orbit Maneuvers of the LeLaR Flight-Agent From Fig.~\ref{fig:maneuver_15.12.2025_09.31-09.49} on 2025-12-15.}} \label{tab:sim_stats_quaternion_list}
\setlength{\tabcolsep}{3pt}
\begin{tabular}{p{43pt}p{34pt}p{34pt}p{34pt}p{34pt}p{26pt}}
\hline
ID (duration)& Comp. & Min ($^\circ$) & Max ($^\circ$)& Mean ($^\circ$) & Std ($^\circ$) \\
\hline
\multirow{3}{*}{M1 (122 s)} & Yaw   & -0.09 &  0.35 &  0.03 &  0.08 \\
                            & Pitch & \textbf{-0.97} &  0.25 &  0.05 &  0.15 \\
                            & Roll  & -0.23 &  \textbf{0.44} & -0.03 &  0.12 \\ \hline
\multirow{3}{*}{M2 (54 s)}  & Yaw   & -0.11 &  0.22 &  0.02 &  0.06 \\
                            & Pitch & \textbf{-0.62} &  0.18 &  0.10 &  0.16 \\
                            & Roll  & -0.43 &  \textbf{0.81} &  0.04 &  0.20 \\ \hline
\multirow{3}{*}{M3 (102 s)} & Yaw   & -0.23 &  \textbf{0.18} &  0.06 &  0.06 \\
                            & Pitch & \textbf{-0.33} &  0.05 & -0.05 &  0.08 \\
                            & Roll  & -0.32 &  0.04 & -0.13 &  0.11 \\ \hline
\multirow{3}{*}{M4 (112 s)} & Yaw   & -0.59 &  0.29 &  0.06 &  0.11 \\
                            & Pitch & -0.44 &  0.15 & -0.02 &  0.11 \\
                            & Roll  & \textbf{-0.80} &  \textbf{0.65} & -0.05 &  0.18 \\ \hline
\multirow{3}{*}{M5 (100 s)} & Yaw   & -0.13 &  \textbf{0.24} &  0.05 &  0.07 \\
                            & Pitch & -0.24 &  0.16 &  0.01 &  0.09 \\
                            & Roll  & \textbf{-0.50} &  0.18 & -0.04 &  0.13 \\ \hline
\multirow{3}{*}{M6 (106 s)} & Yaw   & \textbf{-0.39} &  0.12 & -0.01 &  0.08 \\
                            & Pitch & -0.21 &  \textbf{0.19} &  0.07 &  0.09 \\
                            & Roll  & -0.27 &  0.07 & -0.10 &  0.10 \\ \hline
\end{tabular}
\end{table}

The same list was used for the PD controller of InnoCube. For the first maneuver, the initial attitude was $(0.804, 0.0414, 0.0208, 0.593)$ and the initial satellite body rates were $(0.409, 0.102, 5.57)$~$^\circ${\text{/s}}. Only maneuvers M1 and M5 reached steady-state. Fig.~\ref{fig:pd_maneuver_15.12.2025_22.30-22.48} shows the maneuvers and Table~\ref{tab:pd_stats_15.12.2025_22.30-22.48} presents the metrics for maneuvers M1 and M5.

\Figure[t!](topskip=0pt, botskip=0pt, midskip=0pt)[width=0.99\linewidth]{djebk15.png}
{\textbf{In-orbit flight telemetry for six sequential maneuvers (M1--M6) using the default PD controller on 2025-12-15: (a) attitude Euler angles, (b) commanded RW torques, (c) RW speeds, and (d) satellite body rates. Maneuver intervals with achieved steady-state (Maneuver Start--Steady-State Start; settling-time): M1 (22:30:16--22:31:34; 78~s) and M5 (22:40:16--22:42:30; 134~s). Remaining maneuver start times (Maneuver Start): M2 (22:32:46), M3 (22:35:14), M4 (22:37:46), and M6 (22:42:46).}\label{fig:pd_maneuver_15.12.2025_22.30-22.48}}

\begin{table}[t!]
\caption{\textbf{Steady-State Error Statistics for the Successful Maneuvers M1 and M5 of the PD Controller From Fig.~\ref{fig:pd_maneuver_15.12.2025_22.30-22.48} on 2025-12-15.}}\label{tab:pd_stats_15.12.2025_22.30-22.48}
\setlength{\tabcolsep}{3pt}
\begin{tabular}{p{43pt}p{34pt}p{34pt}p{34pt}p{34pt}p{26pt}}
\hline
ID (duration)& Comp. & Min ($^\circ$) & Max ($^\circ$)& Mean ($^\circ$) & Std ($^\circ$) \\
\hline
\multirow{3}{*}{M1 (72 s)} & Yaw   &  0.05 &  0.74 &  0.23 &  0.18 \\
                    & Pitch &  0.28 &  0.40 &  0.35 &  0.03 \\
                    & Roll  & \textbf{-0.94} &  \textbf{1.67} &  0.72 &  0.84 \\ \hline
\multirow{3}{*}{M5 (16 s)} & Yaw   & -0.12 &  0.26 &  0.13 &  0.12 \\
                    & Pitch & -0.13 &  0.23 &  0.06 &  0.13 \\
                    & Roll  & \textbf{-0.14} &  \textbf{0.52} &  0.18 &  0.21 \\ \hline
\end{tabular}
\end{table}

For the PD controller to attain steady-state during all maneuvers, the threshold had to be relaxed to $2.0^\circ$ per axis. The steady-state metrics for this relaxed threshold are shown in Table~\ref{tab:pd_stats_15.12.2025_22.30-22.48_relaxed}. Consistent with the findings from Section~\ref{sec:repeated_maneuvers}, the PD controller exhibited similar performance to that observed in the repeated maneuvers.

\begin{table}[t!]
\caption{\textbf{Steady-State Error Statistics for PD Maneuvers M1--M6 From Fig.~\ref{fig:pd_maneuver_15.12.2025_22.30-22.48}, Based on a Relaxed Steady-State Threshold of $2^\circ$ Per Axis to Allow the PD Controller to Reach Steady-State During All Maneuvers.}} \label{tab:pd_stats_15.12.2025_22.30-22.48_relaxed}
\setlength{\tabcolsep}{3pt}
\begin{tabular}{p{43pt}p{34pt}p{34pt}p{34pt}p{34pt}p{26pt}}
\hline
ID (duration)& Comp. & Min ($^\circ$) & Max ($^\circ$)& Mean ($^\circ$) & Std ($^\circ$) \\
\hline
\multirow{3}{*}{M1 (112 s)} & Yaw   & \textbf{-1.98} &  0.74 &  0.02 &  0.67 \\
                            & Pitch & -0.10 &  1.19 &  0.30 &  0.20 \\
                            & Roll  & -1.89 &  \textbf{1.67} & -0.14 &  1.28 \\ \hline
\multirow{3}{*}{M2 (110 s)} & Yaw   & \textbf{-1.67} &  0.48 &  0.04 &  0.43 \\
                            & Pitch & -1.19 &  0.84 &  0.02 &  0.41 \\
                            & Roll  & -1.31 &  \textbf{2.01} &  1.44 &  0.58 \\ \hline
\multirow{3}{*}{M3 (88 s)}  & Yaw   &  0.01 &  0.37 &  0.27 &  0.08 \\
                            & Pitch & -0.53 &  0.34 &  0.11 &  0.22 \\
                            & Roll  & \textbf{-2.03} &  \textbf{1.39} & -1.47 &  0.98 \\ \hline
\multirow{3}{*}{M4 (90 s)}  & Yaw   & -0.35 &  0.29 & -0.12 &  0.17 \\
                            & Pitch &  0.12 &  \textbf{0.34} &  0.27 &  0.04 \\
                            & Roll  & \textbf{-5.62} & -1.08 & -2.60 &  1.41 \\ \hline
\multirow{3}{*}{M5 (98 s)}  & Yaw   & -1.19 &  0.55 & -0.08 &  0.44 \\
                            & Pitch & -1.15 &  0.45 &  0.15 &  0.33 \\
                            & Roll  & \textbf{-1.41} &  \textbf{4.06} &  0.73 &  1.42 \\ \hline
\multirow{3}{*}{M6 (80 s)}  & Yaw   & -0.02 &  0.41 &  0.30 &  0.09 \\
                            & Pitch &  0.12 &  \textbf{0.77} &  0.38 &  0.15 \\
                            & Roll  & \textbf{-2.04} & -1.81 & -1.93 &  0.06 \\ \hline
\end{tabular}
\end{table}

Overall, the LeLaR flight-agent reliably performed inertial pointing maneuvers despite Sim2Real discrepancies, with steady-state errors consistently below $1^\circ$ on all axes, meeting the training objective. The agent also adapted to new safety constraints through post-training, and no Safety Cage intervention was triggered during any experiment.

\section{Conclusions}\label{sec:conclusions}
In this work, we presented the design, training methodology, and in-orbit performance of LeLaR, the first AI-based attitude controller successfully demonstrated in space. The results confirm that a controller trained exclusively in simulation can be successfully transferred to flight hardware, validating the feasibility of overcoming the Sim2Real gap for spacecraft attitude control. While previous AI-based attitude control research has primarily been restricted to simulation environments or ground-based testbeds, the successful demonstration of attitude control in orbit is a critical step in confirming the real-world applicability of deep reinforcement learning in aerospace. To our knowledge, this represents the first instance in which an AI-based attitude controller has autonomously oriented a spacecraft in orbit.

Sim2Real discrepancies existed beyond standard modeling uncertainties, including random RW dead times, responsiveness delays, sudden RW speed spikes, and residual magnetic dipole moments exceeding initial assumptions. Despite these, the AI-based controller successfully performed inertial pointing maneuvers in a zero-shot fashion, without prior exposure to the real flight dynamics. This demonstrates that the applied training procedure and reward design enabled the agent to generalize across the Sim2Real gap, which is a critical requirement for autonomous space systems such as deep-space probes, where high-fidelity modeling prior to launch and iterative in-flight fine-tuning are often infeasible. This approach is not intended to replace established control theories, but rather to provide an additional item in the toolbox for addressing spacecraft autonomy.

Although demonstrated on a 3U CubeSat, the proposed methodology is not bound to this class of satellites. Given that the observation space can be covered by the existing sensors, only the simulation model would need to be reconfigured. The fundamental approach remains unchanged.

Regarding computational energy consumption, the neural network inference on the EFR32FG12 microcontroller is negligible, with no meaningful increase in power consumption compared to PD operation. The AI-based controller tends to produce more aggressive actuator commands than the PD controller, which was not a concern for the maneuvers conducted here. If needed, however, this can be mitigated by increasing the smoothness penalty or adding an action penalty in the reward function.

More research is required in automatic optimization of the balance between target achievement and constraint satisfaction, like adhering to rate limits, and on formal verification methods to complement the Safety Cage approach. Our next step is extending the experiment timeframe to demonstrate AI-based momentum management using the magnetorquers in the control loop. Further, we plan to adapt our methodology to other satellites and mission scenarios to validate generalization.

\bibliographystyle{IEEEtran}
\bibliography{bibliography}
\phantomsection
\begin{IEEEbiography}[{\includegraphics[width=1in,height=1.25in,clip,keepaspectratio]{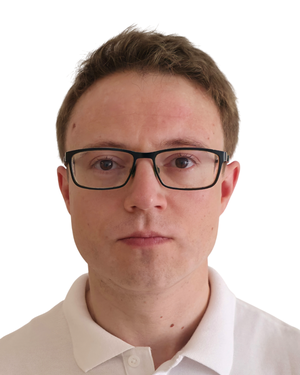}}]{Kirill Djebko}
received the M.S. and Ph.D. (Dr. rer. nat.) degrees in computer science from Julius-Maximilians-Universität Würzburg, Germany, in 2016 and 2020, respectively. 

He is currently a Researcher at the Center for Artificial Intelligence and Data Science (CAIDAS). His research interests include deep reinforcement learning for satellite attitude control, Sim2Real transfer, and the automatic calibration of simulation systems and parameter optimization. He also explores the use of Kolmogorov-Arnold Networks for AI model compression.

Dr. Djebko's current research focuses on AI-based attitude control, specifically investigating the integration of neural networks into real-time control loops and their deployment on resource-constrained hardware.
\end{IEEEbiography}

\begin{IEEEbiography}[{\includegraphics[width=1in,height=1.25in,clip,keepaspectratio]{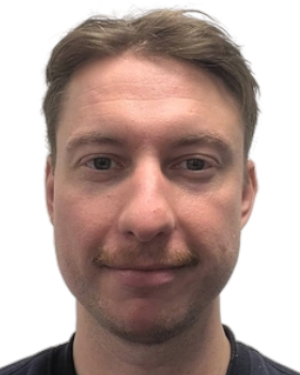}}]{Tom Baumann}
received the M.S. degree in aerospace information technology from Julius-Maximilians-Universität Würzburg, Germany, in 2020.

He is currently a Research Assistant at the Chair of Aerospace Information Technology. He has contributed to several small satellite missions, with a particular focus on the development, integration, and testing of attitude determination and control systems (ADCS). His research interests also include robust system design and the optimization of assembly, integration, and testing (AIT) processes for CubeSats. 

Mr. Baumann's current work centers on mission-specific ADCS hardware-in-the-loop testing and verification strategies for low-Earth-orbit satellites.
\end{IEEEbiography}

\begin{IEEEbiography}[{\includegraphics[width=1in,height=1.25in,clip,keepaspectratio]{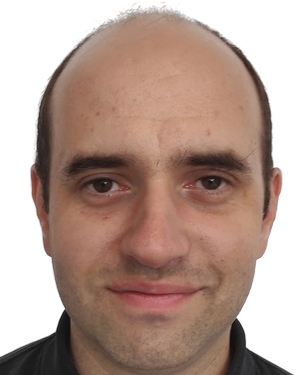}}]{Erik Dilger}
received the M.S. (Dipl.-Inf.) degree in technical computer science from Julius-Maximilians-Universität Würzburg, Germany, in 2013. 

He serves as a Research Assistant at the Chair of Aerospace Information Technology. He has contributed to several research initiatives, including the VIDANA and VaMex projects. Since 2020, he has worked on the InnoCube CubeSat project.

Mr. Dilger's research focuses on the deployment of distributed and redundant systems in space and the development of reliable avionics for small satellite platforms.
\end{IEEEbiography}

\begin{IEEEbiography}[{\includegraphics[width=1in,height=1.25in,clip,keepaspectratio]{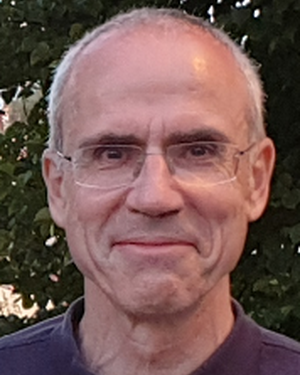}}]{Frank Puppe}
received the M.S. (Dipl.-Inf.) degree in computer science from Bonn University, Germany, the Ph.D. degree (Dr. rer. nat.) from Kaiserslautern University, Germany, and the Habilitation from Karlsruhe University, Germany, in 1983, 1986, and 1991, respectively. 

From 1992 to 2025, he held the Chair of Artificial Intelligence and Knowledge Systems at Julius-Maximilians-Universität Würzburg, Germany. He has authored or coauthored more than 500 publications. His research areas include the combination of large language models and knowledge processing in image and text processing as well as data management and simulation.

Prof. Puppe was honored as a Fellow of the German Informatics Society (Gesellschaft für Informatik) in 2024. He currently serves as a Senior Professor at the Center for Artificial Intelligence and Data Science (CAIDAS).
\end{IEEEbiography}

\begin{IEEEbiography}[{\includegraphics[width=1in,height=1.25in,clip,keepaspectratio]{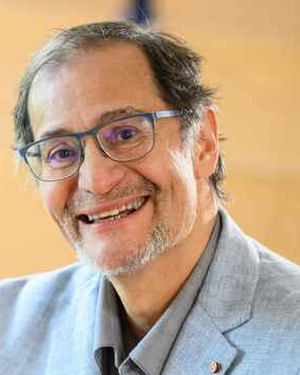}}]{Sergio Montenegro}
received the M.S. (Dipl.-Inf.) and Ph.D. (Dr.-Ing.) degrees in computer science from Technische Universität Berlin, Germany, in 1985 and 1989, respectively.

From 1985 to 2007, he was a Project Manager and Research Coordinator with the Fraunhofer Institute FIRST. Between 2007 and 2010, he served as the Head of the Central Avionics Department at the German Aerospace Center (DLR). He is the creator and maintainer of the Real-time Operating System RODOS. His research interests include avionics for satellites, UAVs, and underwater autonomous vehicles.

Prof. Montenegro has been a Professor and held the Chair of Aerospace Information Technology at Julius-Maximilians-Universität Würzburg, Germany, since 2010. He currently serves as a Senior Professor and continues to lead research in transparent middleware communication for distributed systems.
\end{IEEEbiography}

\EOD

\end{document}